\documentclass[floatfix,rmp,twocolumn,twoside,a4paper]{revtex4}
\usepackage{amsmath}
\usepackage{amsfonts}
\usepackage{amsthm}
\usepackage{reptheorem}

\usepackage{graphicx} 
\usepackage[english]{babel}
\usepackage{makeidx}

\usepackage{booktabs}
\usepackage{cellspace}

\usepackage{hyperref}
\usepackage{xcolor}

\definecolor{darkblue}{RGB}{0,0,192}
\hypersetup{
    colorlinks=true,
    linkcolor=darkblue,
    filecolor=magenta,      
    urlcolor=blue,
    citecolor=blue,
    }
    
\newfont{\chancery}{pzcmi scaled 1200}

\newcommand{\textquote}[3] {
   \begin{quotation}
      {\chancery #1}
      \begin{flushright}
        \normalsize \textsc{#2} \\ {\em #3}
      \end{flushright}
   \end{quotation}
}

\newtheoremstyle{mystyle}
  {6pt}
  {6pt}
  {\it}
  {}
  {\bf}
  {}
  { }
  {{\sc \thmname{#1} \thmnumber{#2}:\thmnote{ #3}}}
\theoremstyle{mystyle}

\theoremstyle{remark}

\bibpunct{(}{)}{,}{n}{}{,}
\addto\captionsenglish{
}

\makeatletter
\renewcommand\tableofcontents{%
    \@starttoc{toc}%
}
\makeatother

\makeindex
        
\begin{document}

\title{DL101 Neural Network Outputs and Loss Functions}
\author{Fernando Berzal \\ \url{berzal@acm.org} }
\affiliation{Department of Computer Science and Artificial Intelligence, University of Granada, Spain}

\maketitle

\tableofcontents
\vspace{6mm}

\vspace{6mm}

\textquote{
I keep saying the sexy job in the next ten years will be statisticians. People think I'm joking, but who would've guessed that computer engineers would've been the sexy job of the 1990s? The ability to take data -to be able to understand it, to process it, to extract value from it, to visualize it, to communicate it- that's going to be a hugely important skill in the next decades... Because now we really do have essentially free and ubiquitous data. So the complimentary scarce factor is the ability to understand that data and extract value from it.
}{Hal R. Varian}{The McKinsey Quarterly, January 2009}

Choosing the right loss function is absolutely central to building effective machine learning models. 

In the case of deep learning models \cite{goodfellow2016,berzal2019a,berzal2019b,aggarwal2023,bishop2024}, the output layer of the neural network can be viewed as a generalized linear model. Given its inputs $(x_1 .. x_n)$, the output of the final layer of the neural network is typically computed as:
$$ y = f \left( \sum_{i=1}^n w_i x_i \right) = f(\vec{w} \cdot \vec{x}) = f(w^\top x)$$
where $f$ is the activation function and $(w_1 .. w_n)$ are the output neuron parameters, a.k.a. weights. 

It is also common that a neuron incorporates a bias $b$, so that its output can  be computed as $f(\vec{w} \cdot \vec{x}+b)$. Assuming an extra input $x_0=1$ with its corresponding weight $w_0=b$ allows us to absorb the bias into the above expression, thus avoiding clutter.

\section{Activation Functions}

The output of an artificial neuron is often just a linear combination of its inputs and its parameters or weights. For such neurons, it is suitable to define its net input (a.k.a. pre-activation) as $z = \vec{w} \cdot \vec{x}$. Such scalar input is transformed by the neuron activation function $f$ to obtain the neuron output $y=f(z)$. Table \ref{activation} summarizes some common activation functions, which are often nonlinear so that a multilayer neural network can act as an universal approximator \cite{cybenko1989, funahashi1989, hornik1989, leshno1993, montufar2014}:

\begin{table*}
\begin{tabular}{SlScScSc} 
\toprule[1.5pt]
 & Activation function & Range & Derivative \\ 
\midrule[1pt]
Linear (identity)  
  & $f(z) = z$ 
  & $(-\infty, \infty) $ 
  & $f'(z) = 1$ \\
Logistic (standard sigmoid)
  & $f(z) = \sigma(z) = \dfrac{1}{1+e^{-z}}$ 
  & $(0,1)$ 
  & $f'(z) = f(z) (1- f(z))$\\
Hyperbolic tangent (tanh)
  & $f(z) = \tanh(z) = \dfrac{e^{z} - e^{-z}}{e^{z} + e^{-z}}$ 
  & $(-1,1)$
  & $f'(z) = 1- f^2(z) $ \\ 
Softmax
  & $ f_i(\vec{z}) = \dfrac{e^{z_i}}{\sum_{j=1}^{K} e^{z_j}}$
  & [0,1] 
  & $\dfrac{\partial y_i}{\partial x_j} = \dfrac{\partial f_i(\vec{z})}{\partial x_j}  = y_i ( \delta_{ij} - y_j)$ \\
Softplus
  & $f(z) = \ln (1+e^z)$
  & $[0, \infty) $
  & $f'(z) = \dfrac{e^z}{1 + e^{z}} = \dfrac{1}{1+e^{-z}} = \sigma(z) $ \\ 
ReLU    
  & $f(z) = \max \{0, z \}$
  & $[0, \infty) $
  & $ 
    f'(z) = H(z) = 
        \begin{cases} 
            1 & \text{if } z > 0 \\ 
            0 & \text{if } z \le 0
        \end{cases}
    $ \\
Leaky ReLU 
  & $ 
    f(z) = 
        \begin{cases} 
            z & \text{if } z > 0 \\ 
            \alpha z & \text{if } z \le 0
        \end{cases}
    $
  & $(-\infty, \infty)$ 
  & $ 
    f'(z) = 
        \begin{cases} 
            1 & \text{if } z > 0 \\ 
            \alpha & \text{if } z \le 0
        \end{cases}
    $ \\
Heaviside step (LTU)
  & $ 
    f(z) = H(z) =
        \begin{cases} 
            1 & \text{if } z > 0 \\ 
            0 & \text{if } z \le 0
        \end{cases}
    $
  & $\{0,1\}$
  & $ 
    f'(z) = \delta(z) =  
        \begin{cases} 
            \infty & \text{if } z = 0 \\ 
            0 & \text{if } z \ne 0
        \end{cases}
    $ \\
\bottomrule[1.5pt]
\end{tabular}
\caption{Common activation functions in neural networks: The activation function determines the output of a neuron, transforming the weighted sum of its inputs into an output signal. For additional activation functions and their variants, check Appendix \ref{appendix-activation-functions}.}
\label{activation}
\end{table*}

\begin{itemize}

\item
The linear or identity function is used mainly in the output layer for regression tasks where the goal is to predict a continuous numerical value. It is also used within specialized network architectures, such as convolutional networks, but in hidden layers it should also be combined with non-linear activation functions (since you could always replace multiple linear layers with a single linear layer, thus loosing the universal approximation properties of multi-layered networks with non-linear activation functions). 

\item
The logistic function ($\sigma$) is a sigmoid function (i.e. S-shaped) that squashes it input into the $[0,1]$ interval, making it suitable for output layers in binary classification problems, when the output can be interpreted as a probability.

\item
The hyperbolic tangent ($\text{tanh}$) function is similar to the sigmoid but maps the input to the $[-1,1]$. It is zero-centered, which often aids in faster convergence during training when compared to the logistic function. It is often used in conjunction with bipolar encoding ${-1,1}$. The hyperbolic tangent and the logistic function are related sigmoid squashing functions: $tanh(z) = 2\sigma(2z) - 1$. 

\item
The softmax function is typically used in the output layer for multi-class classification problems with $K$ classes. It converts a vector of numbers into a probability distribution, where the sum of the probabilities is equal to 1.\footnote{The derivative of the softmax function can be expressed in terms of the Kronecker delta ($\delta_{ij}$) and it results in a Jacobian matrix because softmax is a vector-valued function. The Kronecker delta is $\delta_{ij}=1$ when $i=j$ and $\delta_{ij}=0$ when $i\ne j$. The derivative of the i-th component of the softmax output, $y_i$, with respect to the j-th component of the input, $x_j$, is $y_i(\delta_{ij}-y_j)$.}

\item 
The softplus activation function avoids the main drawback sigmoid functions (logistic and tanh), whose gradients go to zero when their net inputs have large positive or large negative values. For large positive inputs $z \gg 0$, $f(z) \approx \ln(e^z) = z$, so that its gradient is not zero. For large negative inputs $z \ll 0$, $e^z \approx 0$ and $f(z) \approx \ln(1) = 0$, i.e. approaches $0$ but is always positive.

\item 
Rectified linear units ($\text{ReLU}$) are a popular choice for hidden layers in deep learning models. The $\text{ReLU}$ function introduces a single non-linearity and is computationally efficient. The softplus function is a smooth, differentiable approximation of the ReLU function. At $z=0$, the ReLU function has a sharp corner (non-differentiable point), while the softplus function transitions smoothly from near-zero to linear.

\item
The Leaky $\text{ReLU}$ is a variant of the $\text{ReLU}$ function that attempts to solve the ``dying $\text{ReLU}$'' problem by allowing a small, non-zero gradient for negative inputs (typically $\alpha = 0.01$).

\item 
The Heaviside step function was used as the activation function in the linear threshold units used by McCulloch and Pitts \cite{mcculloch1943}. However, it is not used in modern artificial neural networks because its derivative is the Dirac delta function, zero everywhere except at $z=0$, where it is infinitely large. Therefore, it is not suitable for gradient-based learning algorithms.
\end{itemize}

\section{Loss Functions}

We can now return to our original problem, that of choosing the right loss function for our particular problem. Let us consider different tasks and justify which loss functions seem to be the most natural fit:

\begin{itemize}

\item
In \textbf{regression problems}, we predict continuous values, so the loss must measure the distance between predicted and true values. Mean Squared Error (MSE) or Mean Absolute Error (MAE) directly quantify this distance, making them natural choices:
$$ MSE = \frac{1}{n} \sum_{i=1}^{n} (t_i - y_i)^2 $$
$$ MAE = \frac{1}{n} \sum_{i=1}^{n} |t_i - y_i|$$
where $n$ is the number of data points, $t_i$ is the actual (target) value, and $y_i$ is the predicted value.

MSE calculates the average of the squared differences between the predicted values and the actual values. MSE penalizes large errors more heavily because of the square term, which can be useful when you want your model to be especially sensitive to outliers or large deviations.

MAE treats all errors linearly, so it’s more robust / less sensitive to outliers. MAE would be a better choice when you care about median-like behavior rather than mean-like behavior. MAE is also more intuitive to understand as it represents the average error in the same units as the target variable.

\item
\textbf{Binary classification} is about predicting the probabilities of two possible outcomes (i.e. two classes). Binary cross-entropy is the most principled way to compare predicted probabilities with the true labels:
$$H = -\frac{1}{n}\sum_{i=1}^{n}(t_i \cdot \log(y_i) + (1-t_i) \cdot \log(1-y_i))$$
where $n$ is the number of observations, $t_i$ is the actual class label (0 or 1), and $y_i$ is the predicted probability of the $i$-th observation belonging to class 1.

This loss function is applied when the model outputs a probability between 0 and 1. Cross-entropy measures the “distance” between the predicted probability distribution and the true distribution (0 or 1). It quantifies how close the predicted probability is to the actual class label (0 or 1) for each example. A lower binary cross-entropy value indicates a more accurate model. The logarithmic component of the formula heavily penalizes confident but incorrect predictionss, which encourages the model to output well-calibrated probabilities.

\item
In \textbf{multi-class classification problems}, with $K$ different classes ($K>2$), the goal is to predict a probability distribution across $K$ classes (often via the softmax function). Categorical cross-entropy measures how well the predicted distribution matches the true one-hot encoded label:
$$H = - \frac{1}{n} \sum_{i=0}^{n} \sum_{j=0}^{K} t_{ij} \cdot \log(y_{ij})$$
where $K$ is the number of classes, $n$ is the number of observations, $t_{ij}$ is a binary indicator (0 or 1) if class label $j$ is the correct classification for observation $i$, and $y_{ij}$ is the predicted probability of observation $i$ belonging to class $j$.

The categorical cross-entropy ensures the model not only predicts the correct class but also assigns high probability to it relative to other classes. Similar to its binary counterpart, a lower categorical cross-entropy value signifies a better-performing model.

\end{itemize}

As we will see, the above choices of loss functions are far from arbitrary. They are not only natural, but they are also the statistically justified choice for training classification models because minimizing those loss functions is equivalent to maximizing the likelihood of the model predictions.

\section{Statistical Justification}

The formal justification for why mean squared error (MSE) or mean absolute error (MAE), binary cross-entropy (BCE), and categorical cross-entropy (CCE) are the natural loss/error functions for their respective problems comes from the principle of maximum likelihood estimation (MLE), a method for estimating the parameters of an assumed probability distribution, given some observed data. 
In short, choosing a loss function is equivalent to assuming a specific probability distribution for your data. It turns out that these common loss functions are precisely the negative log-likelihoods for standard probability distributions.

\subsection{Maximum Likelihood Estimation}


The MLE principle states that we should choose model parameters that maximize the probability (or likelihood) of observing the training data, i.e. the most likely model that could have produced the observed data. Minimizing the negative log-likelihood (NLL) of the data is the same as maximizing their likelihood. 

The likelihood function measures how "likely" the observed training data are for different values of the model parameters. Assuming independent and identically distributed data (i.i.d.), the likelihood function is the product of the probabilities of each individual observation, given the model parameters:
$$\mathcal{L}(\theta | x_1, x_2, ..., x_n) = P(x_1|\theta) \cdot P(x_2|\theta) \cdot ... \cdot P(x_n|\theta)$$
$$\mathcal{L}(\theta | x_1, x_2, ..., x_n) = \prod_{i=1}^{n} P(x_i|\theta)$$
where $\mathcal{L}(\theta | x_1, ..., x_n)$ is the likelihood of the parameter $\theta$ given the data $x_1, ..., x_n$.

Multiplying many small probabilities can lead to numerical underflow problems (i.e. a rounding error when floating-point numbers near zero are rounded to zero). It is then common to work with the natural logarithm of the likelihood function, called the log-likelihood function. Taking the log turns the product into a sum:
$$\log \mathcal{L}(\theta) = \sum_{i=1}^{n} \log P(x_i|\theta)$$

Since the logarithm is a monotonically increasing function, maximizing the log-likelihood is the same as maximizing the likelihood. Therefore, to find the value of the distribution parameters that maximizes the log-likelihood function of the observed data, you can take the derivative of the log-likelihood function with respect to the model parameters, set it to zero, and solve for the parameters. When closed analytical solutions cannot be found, numerical optimization algorithms are used.

MLE is widely used because its estimates have several desirable properties: consistency (as the sample size grows,  the MLE gets closer to the true value of the parameters), asymptotic normality (with a large enough sample size, the distribution of the MLE is approximately normal), efficiency (for large samples, the MLE has the smallest possible variance among all unbiased estimators), and invariance (if you apply a function to a parameter, the MLE of the transformed parameter is simply the function applied to the original MLE).

The choice of different underlying probability distributions for the observed data leads naturally to different loss functions. Obviously, MLE relies on the assumption that the chosen probability distribution should be a good fit for the data. If that is not the case, your MLE estimates will be inaccurate. Another key assumption is that the data must be independent and identically distributed (i.i.d.), i.e. each data point is drawn from the same underlying distribution and is independent of the others, a requirement that might not always hold in practice.

\subsubsection{Regression and Mean Squared Error (MSE)}

For regression, we want to predict a continuous scalar value $y$ from the input vector $\vec{x}$. We can assume that the target variable $y$ is generated from our model prediction $f(x)$ plus some Gaussian (normal) noise, $\epsilon$.
$$y = f(x) + \epsilon \quad \text{where} \quad \epsilon \sim \mathcal{N}(0, \sigma^2)$$
The noise $\epsilon$ is assumed to have a mean of 0 and a constant variance $\sigma^2$.

This is equivalent to saying the conditional probability of $y$ given $x$ is a Gaussian distribution centered at our model prediction:
$$p(y|x) = \mathcal{N}(y | f(x), \sigma^2) = \frac{1}{\sigma \sqrt{2\pi}} e^{-\frac{1}{2}\left(\frac{y - f(x)}{\sigma}\right)^2}$$

To find the best model parameters, we maximize the likelihood of observing our entire dataset of $n$ i.i.d. points or, equivalently, its log-likelihood:
$$\mathcal{L}(f) = \prod_{i=1}^{n} p(y_i|x_i) = \prod_{i=1}^{n} \mathcal{N}(y_i | f(x_i), \sigma^2)$$
$$\log \mathcal{L}(f) = \sum_{i=1}^{n} \log p(y_i|x_i) = \sum_{i=1}^{n} \log \mathcal{N}(y_i | f(x_i), \sigma^2)$$

Given our assumption of Gaussian noise:
$$\log \mathcal{L}(f) = \sum_{i=1}^{n} {\log\left(\frac{1}{\sigma\sqrt{2\pi}}\right) -\frac{1}{2}\left(\frac{y_i - f(x_i)}{\sigma}\right)^2} $$

$$\log \mathcal{L}(f) = \sum_{i=1}^{n} \log\left(\frac{1}{\sigma\sqrt{2\pi}}\right) - \sum_{i=1}^{n} {\frac{1}{2\sigma^2}\left( y_i - f(x_i)\right)^2}$$

$$\log \mathcal{L}(f) = n \log\left(\frac{1}{\sigma\sqrt{2\pi}}\right) - \frac{1}{2\sigma^2} \sum_{i=1}^{n} {\left( y_i - f(x_i)\right)^2}$$

$$\log \mathcal{L}(f) = -\frac{n}{2} \log \left( 2\pi\sigma^2 \right) - \frac{1}{2\sigma^2} \sum_{i=1}^{n} {\left( y_i - f(x_i)\right)^2}$$

The first term is independent of the model predictions $f(x_i)$, so to maximize the log-likelihood, we must minimize the second term, whose leading factor $1/2\sigma^2$ is also independent of the model predictions. So we just have to minimize
$$ \sum_{i=1}^{n} (y_i - f(x_i))^2$$
which is precisely the sum of squared errors (SSE). Therefore, minimizing the mean squared error MSE (just SSE divided by the constant $n$) is equivalent to performing maximum likelihood estimation under the assumption of Gaussian noise.
In other words, MSE is the negative log-likelihood under Gaussian noise.


Bayes optimality refers to a decision-making principle that minimizes the expected loss, representing the best possible performance for a given task. A model that achieves this lowest possible error rate is called the Bayes optimal classifier (for classification) or Bayes optimal predictor (for regression). It's a theoretical benchmark that sets the performance ceiling; no other model can do better on average. 

In a regression setting, the Bayes optimal predictor, $f^*(x)$, is the function that minimizes this expected squared error. It achieves this by always predicting the conditional expectation of the output given the input:
$$f^*(x) = \mathbb{E}[y | x]$$

Why is the conditional expectation the best possible prediction? Imagine you have a fixed input $x$. There might be a range of possible $y$ values associated with it, each with a certain probability. If you pick any prediction $\hat{y}$, the expected squared error at that point is $\mathbb{E}[(y-\hat{y})^2 | x]$. This expression is minimized when $\hat{y}$ is exactly the average of all possible $y$ values, which is the conditional mean $\mathbb{E}[y | x]$.

From a more formal point of view, we can prove that the conditional mean minimizes the squared loss. Let $Y$ be a random variable with conditional distribution given $X=x$. For any predictor $\hat{y}\in\mathbb{R}$, consider the conditional risk, i.e. the expected loss for a specific decision, given a particular observation:
$$R_2(\hat{y}) = \mathbb{E}\!\left[(Y-\hat{y})^2|X=x\right]$$
$$R_2(\hat{y}) = \mathbb{E}[Y^2|x] - 2\hat{y}\,\mathbb{E}[Y|x] + \hat{y}^2$$

In Bayesian decision theory, the conditional risk is used to evaluate how good a decision-making rule is for each possible data point you might see. The rule that minimizes the conditional risk for every possible observation is called the Bayes estimator or Bayes decision rule. So we differentiate with respect to the prediction $\hat{y}$:
\[
\frac{d}{d\hat{y}}R_2(\hat{y}) = -2\,\mathbb{E}[Y|x] + 2\hat{y}.
\]
Setting the derivative to zero, we obtain that the conditional expectation the best possible prediction:
\[
\hat{y}^* = \mathbb{E}[Y|x].
\]
Given that the second derivative is $2>0$, so $\hat{y}^*$ is the unique minimizer of the MSE. MSE risk is minimized by the conditional mean and learning with the MSE loss function estimates \(\mathbb{E}[Y|X]\).

Therefore, minimizing the MSE is equivalent to finding the Bayes optimal predictor. The remaining error, known as the Bayes error, is caused by the inherent noise or randomness in the data itself. 
Under the assumption of Gaussian noise, minimizing the negative log-likelihood is equivalent to minimizing the mean squared error. Since minimizing MSE gives us the Bayes optimal predictor, it follows that minimizing the NLL under Gaussian noise also leads to the Bayes optimal solution.

\subsubsection{Regression and Mean Absolute Error (MAE)}

If instead of starting from a regression model with additive Gaussian noise,
we assume additive Laplace noise:
$$
y = f(x) + \varepsilon,\quad \varepsilon \sim \text{Laplace}(0,b),
$$
with a Laplace probability density
$$
p(\varepsilon) = \text{Laplace}(0,b) = \frac{1}{2b} e^{-\frac{|\varepsilon|}{b}}
$$ 

The conditional probability of $y$ given $x$ is a Laplace distribution centered at our model prediction:
$$
p(y|x) = \frac{1}{2b} e^{-\frac{|y-f(x)|}{b}},
$$

The log-likelihood of observing our entire dataset of i.i.d. examples is now
$$
\log \mathcal{L}(f) = -n\log(2b) - \frac{1}{b}\sum_{i=1}^n |y_i-f(x_i)|
$$

Maximizing $\mathcal{L}(f)$ is equivalent to minimizing
\[
\sum_{i=1}^n |y_i-\hat{y}_i|,
\]
i.e., the empirical sum of absolute errors. Therefore, MAE is the negative log-likelihood under Laplace noise.

Using Bayesian decision theory, we can check that the conditional median minimizes the absolute loss. For any $\hat{y} \in \mathbb{R}\), the conditional risk is now
$$
R_1(\hat{y}) = \mathbb{E}\!\left[|Y-\hat{y}| \mid X=x\right] = \int |y-\hat{y}|\, dF_{Y\mid x}(y).
$$

The conditional risk function is $R_1(\hat{y})$, the mean absolute error (MAE), and our goal is to find the prediction $\hat{y}$ that minimizes it. The problem is that the absolute value function, $|z|$, is not differentiable at $z=0$. Since we cannot just take the standard derivative of $R_1(\hat{y})$ with respect to $\hat{y}$ and set it to zero, we find the subgradient\footnote{A subgradient is a way to generalize the concept of a derivative (or gradient) to functions that aren't smooth or differentiable everywhere. For a convex function, instead of a single tangent line at a point, you might have a whole set of lines that lie below the function. Any of these lines' slopes is a subgradient. A subgradient allows us to optimize a function even when the original function is not perfectly differentiable.} of the risk function $R_1(\hat{y})$. We split the risk integral into two parts, based on where $y-\hat{y}$ is positive or negative:
$$R_1(\hat{y}) = \int_{-\infty}^{\hat{y}} (\hat{y}-y)\, dF_{Y\mid x}(y) + \int_{\hat{y}}^{\infty} (y-\hat{y})\, dF_{Y\mid x}(y)$$

Now, we can differentiate with respect to $\hat{y}$ using the Leibniz rule for differentiating under the integral sign:
$$\partial R_1(\hat{y}) = \int_{-\infty}^{\hat{y}} 1 \, dF_{Y\mid x}(y) - \int_{\hat{y}}^{\infty} 1 \, dF_{Y\mid x}(y)$$

The integrals of the probability distribution function are, by definition, probabilities:
$$\int_{-\infty}^{\hat{y}} dF_{Y\mid x}(y) = P(Y \le \hat{y} \mid X=x) = F_{Y\mid x}(\hat{y})$$
$$\int_{\hat{y}}^{\infty} dF_{Y\mid x}(y) = P(Y > \hat{y} \mid X=x) = 1 - F_{Y\mid x}(\hat{y})$$

Substituting these back we obtain a subgradient  with respect to $\hat{y}$ is
$$
\partial R_1(\hat{y}) = F_{Y\mid x}(\hat{y}) - (1 - F_{Y\mid x}(\hat{y})) = 2F_{Y\mid x}(\hat{y}) - 1,
$$
where \(F_{Y\mid x}\) is the conditional cdf. 

Any $\hat{y}$ with $0\in \partial R_1(a)$ satisfies $F_{Y\mid x}(a)=\frac{1}{2}$, i.e., $\hat{y}$ is a conditional median, the point where the cumulative distribution function (CDF) equals 0.5. Thus, minimizing the expected absolute loss yields the conditional median.

\subsubsection{Binary Classification and Binary Cross-Entropy}

For binary classification problems, the target $y$ is either 0 or 1, $y \in \{0,1\}$. We can assume that the outcome follows a Bernoulli distribution. The model, with a sigmoid output $\hat{y}$, predicts the probability of the outcome being 1. The  classifier outputs $\hat{y}\in(0,1)$:
$$P(y=1|x) = \hat{y}$$
$$P(y=0|x) = 1 - \hat{y}$$

Both expressions can be combined into a more succinct single expression, the Bernoulli likelihood for a single observation:
$$P(y|x) = \hat{y}^y (1-\hat{y})^{1-y}$$

Assuming i.i.d. examples, as always, the likelihood for the whole training dataset is:
$$\mathcal{L} = \prod_{i=1}^{n} P(y_i|x_i) = \prod_{i=1}^{n} \hat{y}_i^{y_i} (1-\hat{y}_i)^{1-y_i}$$

Taking the logarithm gives the log-Likelihood:
$$\log \mathcal{L} = \sum_{i=1}^{n} \left[ y_i \log(\hat{y}_i) + (1-y_i) \log(1-\hat{y}_i) \right]$$

Since our goal is maximizing this log-likelihood, we can also minimize the negative of this quantity. The averaged (per example) negative log-likelihood is precisely the binary cross-entropy:
$$BCE = - \frac{1}{n} \sum_{i=1}^{n} \left[ y_i \log(\hat{y}_i) + (1-y_i) \log(1-\hat{y}_i) \right]$$

Over the training examples, minimizing the average BCE is equivalent to the maximum likelihood estimate for a Bernoulli model: minimizing the binary cross-entropy is identical to maximizing the log-likelihood of the data under the assumption of a Bernoulli distribution.

True probabilities minimize the expected log loss (i.e. the binary cross-entropy). 
Given $x$, for any candidate $\hat{y}\in(0,1)$, the conditional risk is
$$
R_{\text{BCE}}(\hat{y}) = \mathbb{E}\!\left[-Y\log \hat{y} - (1-Y)\log(1-\hat{y}) \mid X=x\right]
$$
$$
R_{\text{BCE}}(\hat{y}) = -y\log \hat{y} - (1-y)\log(1-\hat{y}),
$$

We minimize such conditional risk by differentiation:
$$
\frac{d}{d\hat{y}} R_{\text{BCE}}(\hat{y}) = -\frac{y}{\hat{y}} + \frac{1-y}{1-\hat{y}}
$$

Setting the derivative to zero:
$$
-\frac{y}{\hat{y}} + \frac{1-y}{1-\hat{y}} = 0 \quad \Rightarrow \quad (1-y)\hat{y} = y(1-\hat{y}) \quad \Rightarrow \quad \hat{y} = y
$$

The second derivative is
$$
\frac{d^2}{d\hat{y}^2} R_{\text{BCE}}(\hat{y}) = \frac{y}{\hat{y}^2} + \frac{1-y}{(1-\hat{y})^2} > 0
$$
shows that the conditional risk is strictly convex, so $\hat{y}=y$ is its unique minimum. Therefore, the log loss is a strictly proper scoring rule: it is minimized by the true conditional probability.


\subsubsection{Multi-Class Classification and Categorical Cross-Entropy}

For multi-class classification, the target $y$ belongs to one of $K$ classes.
We can now assume that the outcome follows a multinomial distribution, also known as categorical, multinoulli, or generalized Bernoulli distribution. The true label $y_i$ is represented as by one-hot encoded vector and the model softmax output $\hat{y}_i$ is a probability distribution, i.e. a vector of probabilities for each of the $K$ classes:
$$\hat{y}_i = [p_{i1}, p_{i2}, ..., p_{iK}]$$

The probability of observing the one-hot vector $y_i$ is:
$$P(y_i|x_i) = \prod_{k=1}^{K} p_{ik}^{y_{ik}}$$
Since $y_{ik}$ is 1 only for the true class and 0 otherwise, this product simply picks the predicted probability for the correct class.

The likelihood for the whole training dataset is:
$$\mathcal{L} = \prod_{i=1}^{n} \prod_{k=1}^{K} p_{ik}^{y_{ik}}$$

Taking the logarithm gives the log-likelihood:
$$\log \mathcal{L}  = \sum_{i=1}^{n} \sum_{k=1}^{K} y_{ik} \log(p_{ik})$$

Maximixing this log-likelihood is equivalent to minimizing its negative, the negative log-likelihood. In this case, the negative log-likelihood is just the categorical cross-Entropy loss:
$$CCE = - \sum_{i=1}^{n} \sum_{k=1}^{K} y_{ik} \log(p_{ik})$$

Therefore, minimizing the categorical cross-entropy is formally equivalent to maximizing the log-likelihood of the data under the assumption of a multinomial distribution: the categorical cross-entropy is the multinomial negative log-likelihood.

Let us now check how the true distribution minimizes the expected cross-entropy.
Given $x$, let $y=(y_1,\dots,y_K)\) be the true class distribution and $\hat{y}$ any candidate model output. The conditional risk is now
$$
R_{\text{CCE}}(\hat{y}) = \mathbb{E}\!\left[-\sum_{k=1}^K Y_k \log \hat{y}_k \mid X=x\right]
= -\sum_{k=1}^K y_k \log \hat{y}_k.
$$
This is the cross-entropy $H(y,\hat{y})$, since
$$
-\sum_{k} y_k \log \hat{y}_k = -\sum_{k} y_k \log y_k + \sum_{k} y_k \log \frac{y_k}{\hat{y}_k}
$$
$$
H(y,\hat{y}) = H(y) + \mathrm{KL}(y\|\hat{y})
$$

Given that $H(y)$ is constant with respect to $\hat{y}$ and $\mathrm{KL}(y\|\hat{y}) \ge 0$ with equality if and only if $\hat{y}=y$, the unique minimizer of the conditional risk is \(\hat{y}=y\). Therefore, categorical cross-entropy is a strictly proper scoring rule: its expected value is uniquely minimized when the predicted probability distribution exactly matches the true (or empirical) probability distribution.


\subsubsection{Summary}

\begin{table*}[!t]
\begin{tabular}{SlScScScSc} 
\toprule[1.5pt]
Problem type & Output activation & Loss function & Distribution & MLE estimate \\ 
\midrule[1pt]
Regression
  & Linear
  & MSE
  & Gaussian
  & Mean \\
Regression
  & Linear
  & MAE
  & Laplace
  & Median \\
Regression
  & Linear
  & $\log$
  & Pareto
  & Mode \\
Classification: Binary 
  & Logistic sigmoid
  & BCE
  & Bernoulli
  & Conditional probability \\
Classification: Binary  (bipolar)
  & Hyperbolic tangent
  & BCE
  & Bernoulli
  & Conditional probability \\
Classification: Multi-class 
  & Softmax
  & CCE
  & Multinomial
  & Class probability distribution \\
\bottomrule[1.5pt]
\end{tabular}
\caption{Common loss functions and their MLE justification.}
\label{mle}
\end{table*}
We have seen that choosing loss functions is not arbitrary. Common loss functions arise as the negative log-likelihood of natural probabilistic models and yield Bayes-optimal estimators under common noise assumptions (see Table \ref{mle}). 

\begin{itemize}

\item
For regression problems, MSE is justified by Gaussian MLE and Bayes-optimality for the mean; whereas MAE is the result of Laplace MLE and Bayes-optimality for the median.

\item
For binary and multi-class classification, cross-entropy is the negative log-likelihood of Bernoulli and multinomial models, respectively, and is a strictly proper scoring rule, uniquely minimized by the true conditional probabilities. Properness ensures calibrated probability estimation.
\end{itemize}


With respect to their robustness:

\begin{itemize}

\item
MSE is optimal under homoscedastic Gaussian noise and targets the mean. It is sensitive to outliers because Gaussian tails penalize large deviations quadratically.

\item
MAE is optimal under Laplace noise and targets the median. Heavier tails yield linear penalties and robustness to outliers.    

\item
Cross-entropy arises from discrete likelihoods. Its strict convexity in predicted probabilities penalizes overconfident misclassifications and aligns with maximum likelihood estimation for categorical outcomes.

\end{itemize}

The loss function used in training deep learning models is the negative log-likelihood of the chosen distribution. That is the reason why the ``right'' loss function depends on the prediction task.

\subsection{Generalized Linear Models}

The output layer of many deep neural networks can be seen as a generalized linear model (GLM), where the activation function of the output neurons corresponds to the inverse of a canonical link function and the error (or loss) function is derived from the probability distribution assumed by the GLM. This connection provides a statistical foundation for the design of neural network output layers for various machine learning tasks, since the loss functions discussed (MSE, MAE, cross-entropy) are the negative log-likelihoods of GLMs under different assumptions about the distribution of the response variable.



Generalized linear models (GLMs) extend the principles of linear regression to response variables that are not normally distributed. A GLM is composed of three key elements:

\begin{itemize}
\item
A random component: The probability distribution of the response variable \(Y\), taken from the exponential family, a broad class of distributions that includes the Gaussian, Bernoulli, binomial, multinomial, exponential, gamma, beta, chi-squared, geometric, Dirichlet, Whishart, and Poisson distributions.
\item
A systematic component: A linear predictor, \(\eta = X\beta\), which is a linear combination of the input features. The model parameters $\beta$ would be the output layer weights $w$ in a deep learning context.
\item
A link function: A function \(g(\cdot)\) that connects the expected value of the random component (e.g. the mean of \(Y\)) to the systematic component, i.e. the linear predictor (e.g. \(g(\mu) = \eta\)).

\end{itemize}


A distribution is in the exponential family if its probability mass/density function can be written in the form:
$$p(y | \eta) = h(y) e^{\eta T(y) - A(\eta)}$$
where $y$ is the outcome variable, $\eta$ (eta) is the natural parameter of the distribution, $T(y)$ is the sufficient statistic (often, just $y$), $h(y)>0$ is the base measure, and $A(\eta)$ is the log-partition of the particular probability distribution.


Canonical link functions are a specific type of link function that have desirable mathematical properties, often leading to more stable and efficient model fitting. Each probability distribution in the exponential family has a unique canonical link function.
In a generalized linear model (GLM), the canonical link function, $g(.)$, is the specific function that maps the expected value of the outcome, $\mu = \mathbb{E}[Y]$, to the natural/canonical parameter, $\eta$, i.e. $\eta = g(\mu)$.
When $\eta(\theta) = \theta$, the exponential family is said to be in canonical form. In the special case that $\eta(\theta) = \theta$ and $T(y)=y$, the distribution is in the natural exponential family (NEF), a special case of the exponential family.

This choice is ``canonical'' or natural because it directly links the model linear predictor to the distribution natural parameter. Given the canonical link function, an alternative, equivalent form for members of the exponential family of probability distributions is
$$p(y | \eta,s) = \frac{1}{s}h\left(\frac{y}{s}\right) g(\eta) e^{\frac{\eta y}{s}}$$
taking into account that, when $f(x)$ is a normalized density function, $\frac{1}{s} f\left( \frac{y}{s} \right)$ is also a normalized density function.

Therefore, the function $g(\eta)$ can be interpreted as the coefficient that ensures that the distribution is normalized:
$$ g(\eta) \text{ s.t. } \int p(y | \eta,s) dy = 1$$
which can be used to compute the conditional expectation of $y$, $\mathbb{E}[y]$
$$ \mathbb{E}[y|\eta,s] = \int p(y | \eta,s) y dy$$
by computing the gradient of the above equality with respect to the natural parameter $\eta$:
$$ \nabla_\eta \int p(y | \eta,s) dy = \nabla_\eta 1 = 0$$
$$ \nabla_\eta g(\eta) \int \frac{1}{s}h\left(\frac{y}{s}\right) g(\eta) e^{\frac{\eta y}{s}} dy = 0$$

\begin{align*}
& \nabla_\eta g(\eta) \int \frac{1}{s} h\left(\frac{y}{s}\right) e^{\frac{\eta y}{s}} dy 
\\
& + g(\eta) \int \frac{1}{s} h\left(\frac{y}{s}\right) e^{\frac{\eta y}{s}} \left( \frac{y}{s} \right) dy = 0
\end{align*}

\begin{align*}
& \frac{\nabla_\eta g(\eta)}{g(\eta)} \int \frac{1}{s} h\left(\frac{y}{s}\right) g(\eta) e^{\frac{\eta y}{s}} dy 
\\
& + \frac{1}{s} \int \frac{1}{s} h\left(\frac{y}{s}\right) g(\eta)  e^{\frac{\eta y}{s}} y  dy = 0
\end{align*}

$$
\frac{\nabla_\eta g(\eta)}{g(\eta)} \int p(y | \eta,s) dy 
+ \frac{1}{s} \int p(y | \eta,s) y  dy = 0
$$

$$
\frac{\nabla_\eta g(\eta)}{g(\eta)} \cdot 1
+ \frac{1}{s} \mathbb{E}[y|\eta,s] = 0
$$
so
$$
\mathbb{E}[y|\eta,s] = - s \frac{\nabla_\eta g(\eta)}{g(\eta)} = - s \frac{d}{d \eta} \log g(\eta)
$$


The log likelihood for this model, in terms of $\eta$, is
$$\log p(y | \eta) = \sum_{i=1}^n \log \left[ \frac{1}{s}h\left(\frac{y_i}{s}\right) g(\eta_i) e^{\frac{\eta_i y_i}{s}} \right]$$
$$\log p(y | \eta) = \sum_{i=1}^n \left[ -\log s + \log h\left(\frac{y_i}{s}\right) + \log g(\eta_i) + {\frac{\eta_i y_i}{s}} \right]$$

Assuming that all examples share a common scale parameter, $s$, and $h(y_i/s)$ are independent of $i$ and the $\log h(y_i/s)$ term can be considered constant. The derivative of the log likelihood with respect to the model parameters is then
$$\nabla_w \log p(y | \eta) = \sum_{i=1}^n \left[ \frac{d}{d\eta_i} \log g(\eta_i) + {\frac{y_i}{s}} \right] \frac{d \eta_i}{d y_i} \frac{d y_i}{d z_i} \nabla_w z_i$$
where $z_i = w_i x_i$.

Given our expression for the conditional mean of $y$
$$ \frac{d}{d\eta_i} \log g(\eta_i) = - \frac{1}{s} \mathbb{E}[y_i|\eta] = - \frac{\hat{y_i}}{s}$$
Therefore
$$\nabla_w \log p(y | \eta) = \sum_{i=1}^n \frac{1}{s} \left[ y_i - \hat{y_i} \right] \psi'(\hat{y_i}) f'(z_i) x_i$$
where 
$\hat{y_i} = f(z_i)$, 
$f'(z_i)=\frac{d y_i}{d z_i}$, 
$\psi'(\hat{y_i})=\frac{d \eta_i}{d y_i}$, and 
$x_i= \nabla_w z_i = \nabla_w w_i x_i$.

That expression can be simplified if we choose a particular link function $f^{-1}(\hat{y})=\psi(\hat{y})$. In other words, the activation function used in the output layer of a neural network is the inverse of the canonical link function, $g^{-1}(\eta)$.

Given that $z=f^{-1}(\hat{y})$, $z=\psi(\hat{y})$. Since $f'(\psi) \psi'(\hat{y})=1$, $f'(z) \psi'(\hat{y})=1$. Therefore, the gradient of the log likelihood reduces to 
$$\nabla_w \log p(y | \eta) = \frac{1}{s} \sum_{i=1}^n  \left[ y_i - \hat{y_i} \right] x_i$$


It comes as no surprise that the derivative of the log likelihood with respect to the parameter vector, used by gradient-descent algorithms to train neural networks, always results in contributions to the error function of the form $\left[ y_i - \hat{y_i} \right] x_i$ when the natural pairs of activation and loss functions are chosen (linear/MSE, sigmoid/BCE, softmax/CCE). The formal connection between the loss functions derived from maximum likelihood estimation and canonical link functions lies in the exponential family of distributions.

\subsubsection{Regression: Gaussian GLM}

In a regression problem, the goal is to predict a continuous value. Linear regression is a GLM with Gaussian errors and the identity link. MSE is its natural loss. From a GLM perspective:

\begin{itemize}

\item
Random component: 
The response variable is assumed to follow a Gaussian (normal) distribution, $Y \sim \mathcal{N}(\mu, \sigma^2)$.

\item
Canonical link function: 
For a Gaussian distribution, the expected value is its mean, so the natural parameter is $\eta = \mathbb{E}[Y] = \mu$. The canonical link function connects the expected value to the natural parameter, so, for the Gaussian distribution, the canonical link function is the function $g(\mu) = \eta$. Since both $\eta$ and $\mu$ are the same, the canonical link function is the identity function $g(\mu) = \mu$.  This means the expected value of the response is directly modeled by the linear predictor: \(\mu = X\beta\).

\end{itemize}

As shown previously, assuming a Gaussian distribution for the data leads directly to the MSE loss function via maximum likelihood estimation: minimizing the MSE is mathematically equivalent to maximizing the log-likelihood of the data under a Gaussian distribution.

The canonical link for the Gaussian distribution is the identity function. The inverse link, $g^{-1}(\eta)$, which dictates the neural network output layer activation, is also the identity function (i.e., a linear layer). This formally connects the linear output layer and MSE loss function to the canonical link of the underlying Gaussian probability assumption.

In summary, a neural network with a linear output layer trained with MSE is performing a form of non-linear regression, where the hidden layers learn a complex transformation of the input features, and the output layer acts as a simple linear regression on these learned features.

\subsubsection{Regression: Laplace GLM}

Linear regression in terms of MAE is a GLM with Laplace errors and the identity link. The GLM goal is minimizing the average absolute difference between predicted and actual values rather than the more common mean squared error:

\begin{itemize}

\item
Random component: 
The response variable is assumed to follow a Laplace distribution, $Y \sim \text{Laplace}(\mu, b)$, with mean $\mu$ and scale parameter $b$.

\item
Canonical link function: 
For a Laplace distribution, the expected value is its mean, so its natural parameter is $\eta = \mathbb{E}[Y] = \mu$. For the Laplace distribution, the canonical link function is the function $g(\mu) = \eta$. Since both $\eta$ and $\mu$ are the same, the canonical link function is the identity function $g(\mu) = \mu$.  This means the expected value of the response is directly modeled by the linear predictor: \(\mu = X\beta\).

\end{itemize}

Assuming a Laplace distribution for the data leads directly to the MAE loss function via maximum likelihood estimation: minimizing the MAE is mathematically equivalent to maximizing the log-likelihood of the data under a Laplace distribution.

As in the Gaussian case, since the canonical link is the identity function, the inverse link, $g^{-1}(\eta)$, which dictates the neural network output layer activation, is also the identity function (i.e., a linear layer). This formally connects the linear output layer and MAE loss function to the canonical link of the underlying Laplace probability assumption.

\subsubsection{Binary Classification: Bernoulli GLM}

For binary classification, the objective is to predict one of two possible classes. Logistic regression is a GLM with Bernoulli distribution and logit link. Binary cross-entropy is its natural loss. From a GLM perspective:

\begin{itemize}
\item
Random component: The response variable is assumed to follow a Bernoulli distribution, \(Y \sim \text{Bernoulli}(p)\). The expected value of a Bernoulli variable is its probability of success, so $\mu = \mathbb{E}[Y] = p$.

\item
Canonical link function: The canonical link function for the Bernoulli distribution is the logit function: $\eta = logit(p) = \log\left(\frac{p}{1-p}\right)$, where $p$ is the probability of success (i.e., the probability of $y=1$). The logit transforms probabilities into log odds, given that an event odds is just the ratio of the probability of the event occurring to the probability of not occurring, $p/(1-p)$. 
\end{itemize}

As shown in the previous Section, assuming a Bernoulli distribution leads directly to the binary cross-entropy loss function (also known as log loss): Minimizing the binary cross-entropy is equivalent to maximizing the log-likelihood of the data under a Bernoulli distribution.

The inverse link, $g^{-1}(\eta)$, gives us the required activation function:
$$p = \text{logit}^{-1}(\eta) = \frac{1}{1 + e^{-\eta}}$$
The inverse of the logit function is the logistic (or standard sigmoid) function, which maps the linear predictor to a probability between 0 and 1.
This formally connects the sigmoid activation function and the binary cross-entropy loss to the canonical link of the Bernoulli distribution.

A neural network with a single sigmoid output unit trained with binary cross-entropy loss is essentially learning a logistic regression model on a transformed feature space created by its hidden layers.

\subsubsection{Multi-Class Classification: Multinomial GLM}

In multi-class classification, the task is to predict one of several mutually exclusive classes. Multinomial logistic regression, also known as softmax regression, is a GLM with multinomial distribution and generalized logit link. Categorical cross-entropy is its natural loss.

\begin{itemize}
\item
Random component:
The response variable is assumed to follow a multinomial/categorical  distribution,  \(Y \sim \text{Multinomial}(1, p)\). Its expected value is the vector of class probabilities, $\mu = \mathbb{E}[Y] = [p_1, \dots, p_K]$.

\item
Canonical link function:  
The canonical link connects the vector of probabilities to the natural parameters. For $K$ classes, there are $K-1$ natural parameters, $\eta_k = \log\left(\frac{p_k}{p_K}\right)$. This is the generalized logit function.

\end{itemize}

As shown before, assuming a categorical/multinomial distribution leads directly to the categorical cross-entropy loss function: Minimizing the categorical cross-entropy loss is equivalent to maximizing the log-likelihood of the data under a categorical/multinomial distribution.

The inverse of the generalized logit link function, $g^{-1}(\eta)$, which maps the linear outputs back to probabilities, is the softmax function, which takes a vector of arbitrary real numbers and transforms them into a probability distribution over the problem classes:
$$p_k = \frac{e^{\eta_k}}{\sum_{j=1}^{K} e^{\eta_j}}$$

A neural network with a softmax output layer trained with categorical cross-entropy loss is effectively a multinomial logistic regression classifier (also known as softmax classifier) that operates on the high-level features extracted by the preceding layers of the network.

\subsubsection{Summary}

\begin{table*}[t!]
\begin{tabular}{SlScScScScSc} 
\toprule[1.5pt]
Problem type & GLM & GLM distribution & Output activation & Link function  & Loss function \\ 
\midrule[1pt]
Regression
  & Linear regression
  & Gaussian  
  & Linear
  & Identity
  & MSE \\
Regression
  & Linear regression
  & Laplace   
  & Linear
  & Identity
  & MAE\\
Binary classification
  & Logistic regression
  & Bernoulli 
  & Logistic sigmoid
  & Logit
  & BCE\\
Binary classification (bipolar)
  & Logistic regression
  & Bernoulli 
  & Hyperbolic tangent
  & Logit
  & BCE\\
Multi-class classification
  & Softmax regression
  & Multinomial 
  & Softmax
  & Generalized logit
  & CCE\\
\bottomrule[1.5pt]
\end{tabular}
\caption{Common loss functions for different problem types and their GLM justification.}
\label{glm}
\end{table*}

The choice of activation function (linear, sigmoid, softmax) is the inverse of the canonical link function for the assumed data distribution, and the choice of loss function (MSE, cross-entropy) is the negative log-likelihood of that same distribution. This theoretical foundation provides a cohesive, statistically-grounded framework for designing neural network output layers.

\section{Additional Situations}

\subsection{Binary Classification with Bipolar Encoding}

The loss function for binary classification problems was obtained for logistic activation functions, where the logistic output estimates the probability of the example belonging to the positive class, $p(y=1\mid x)$. Training data was then coded so that $y=1$ indicated that the example belonged to the positive class, whereas $y=0$ indicated that that example belonged to the negative class.

However, a bipolar encoding of output classes is also common in practice: $y=+ 1$ for the positive class, $y=-1$ for the negative class. In that case, we are just scaling and shifting our binary output:
$$ y_{bipolar} = 2 y - 1$$
so that 
$$ y = \frac{y_{bipolar}+1}{2}$$

The bipolar loss function can be derived from the binary cross-entropy log likelihood:
$$\log \mathcal{L} = \sum_{i=1}^{n} \left[ y_i \log(\hat{y}_i) + (1-y_i) \log(1-\hat{y}_i) \right]$$
just by applying the above transformation both to the target variable $y_i$ and the model estimation $\hat{y}_i$.

\begin{align*}
\log \mathcal{L} = \sum_{i=1}^{n}  
& \frac{y_i + 1}{2} \log \left( \frac{\hat{y}_i + 1}{2} \right) \\
& + \left( 1 - \frac{y_i+1}{2} \right) \log \left( 1 - \frac{\hat{y}_i+1}{2} \right) 
\end{align*}

\begin{align*}
\log \mathcal{L} = \frac{1}{2} \sum_{i=1}^{n} 
& (y_i+1) \log \left( \hat{y}_i + 1 \right) \\
& + \left( 1 - y_i \right) \log \left( 1 - \hat{y} \right) 
- n \log 2
\end{align*}

The last term $n\log 2$ does not affect the log likelihood maximization, since it is independent of the model parameters, so it can be dropped.

The activation function that matches the above BCE-like minimization problem can be obtained by applying the linear transformation to the logistic function, which is the  activation function of choice for binary classification problems:
\begin{align*}
y(z) & = 2 \sigma(z) -1 \\
     & = \frac{2}{1+e^{-z}} -1 \\
     & = \frac{1- e^{-z}}{1+e^{-z}}  \\
     & = \frac{e^{z/2}}{e^{z/2}} \frac{1- e^{-z}}{1+e^{-z}} \\
     & = \frac{e^{z/2} - e^{-z/2}}{e^{z/2}+e^{-z/2}} \\
     & = \tanh (z/2)
\end{align*}

Therefore, the hyperbolic tangent activation function is the suitable activation function to be used for solving binary classification problems using bipolar encoding when we want to minimize the binary cross-entropy (i.e. the negative log-likelihood of the data under the assumption of a Bernoulli distribution).

\subsection{Regression of Positive Values}

In many real-world applications, regression models should always return a positive value for predicting prices, counts, durations, or intensities (e.g. waiting times, insurance claims, rainfall amounts, energy consumption...). The choice of activation/loss pair depends on the distribution of the target variable and the constraints you want to enforce. Let us consider several alternatives:

\begin{itemize}

\item
The quick and dirty solution is resorting to conventional regression loss functions. Since they do not enforce positivity, you can constraint the output using an activation function that ensures that the network output is in the correct domain. ReLU and softplus ensure non-negative values.


ReLU can be seen as a piecewise linear approximation to the log link role (the canonical link for a Poisson likelihood). ReLU enforces non‑negativity, but grows linearly instead of exponentially, which might prevent numerical problems. Even though ReLU is not the exact MLE link, it is a computationally efficient surrogate that enforces the positivity constraint.


The softplus activation function also ensures positivity. Softplus is a smooth approximation to ReLU. Softplus behaves like $e^z$ for large negative $z$ (small positive outputs) and like $z$ for large positive $z$ (linear growth, like ReLU). It also has a stronger tie to likelihood theory. For distributions requiring strictly positive parameters (e.g. variance in Gaussian, scale in Gamma, rate in Poisson), the canonical MLE link is the log link ($\theta = e^z$).
Softplus can be loosely interpreted as a numerically stable approximation to the exponential link used in MLE, avoiding overflow problems while still enforcing positivity.

\item 
A popular strategy is training the network to predict $\log \hat{y}$ instead of $\hat{y}$. You can use the MSE loss function on $\log y$ vs. $\log \hat{y}$.
The log transformation naturally enforces positivity: the final output is $e^{\hat{\nu}}$, where $\nu$ is the model output. It also makes the model target multiplicative errors, since $y = f(x) \cdot\epsilon$ becomes $\log y = \log f(x) + \epsilon$.

In practice, therefore, a model is trained to predict $\hat{\nu}=\log \hat{y}$ and the final prediction is $\hat{y}=e^{\hat{\nu}}$. 

Optionally, you can perform bias correction: If errors on log scale are approximately Gaussian with variance $\sigma^2$, an unbiased mean estimate is $\hat{\mu} = e^{\hat{\nu}+\sigma^2/2}$, where the variance $\sigma^2$ itself can be estimated from residuals.

Since the output of the model is transformed using an exponential function, it does not need to be constrained and a simple linear output layer can be used with the logarithmic transformation.

\item
The Gamma distribution is a distribution of the exponential family set of probability distributions that generalizes the exponential, Erlang, and chi-squared distributions. 
$$ p(x) = \frac{\lambda^\alpha}{ \Gamma(\alpha)} x^{\alpha-1} e^{-\lambda x}$$
where $\alpha$ is a shape parameter and $\lambda=1/\theta$ is a rate parameter (an equivalent parameterization uses the $\alpha$ shape and $\theta$ scale parameters). Its mean is $\mu = \alpha/\lambda = \alpha \theta$ and its variance is $\sigma^2 = \alpha/\lambda^2 = \alpha \theta^2$. 

The log likelihood of the Gamma distribution (for a single example) is
$$ \log p(x) = \alpha \log \lambda - \log \Gamma(\alpha) + (\alpha-1) \log x - \lambda x$$

Minimizing the negative log likelihood of the Gamma distribution, a model can learn to output the two positive parameters of the Gamma distribution using softplus activation functions (to ensure positivity). For stability, a small constant $\varepsilon \approx 10^{-3}$ is added to the softplus output (to avoid zero parameters). The model can then be used to predict values (using the mean of the distribution $\mu = \alpha/\lambda$) or sample from the distribution to provide uncertainty estimates.

We could directly predict the mean $\hat{y} = \hat{\mu}$ by reparameterizing $\alpha = \lambda \mu$:
$$ p(x) = \frac{x^{\lambda\mu -1} \lambda^{\lambda\mu} e^{-\lambda x}}{ \Gamma(\lambda\mu)}$$
and maximizing the following log likelihood:
\begin{align*} 
\log \mathcal{L} 
& = \sum_{i=1}^n \left( \lambda\hat{y_i} \log \lambda + (\lambda\hat{y_i} -1)\log y_i - \lambda y_i - \log \Gamma(\lambda\hat{y_i})  \right) \\
& = \sum_{i=1}^n \left( \lambda\hat{y_i} ( \log \lambda + \log y_i) - \log y_i - \lambda y_i - \log \Gamma(\lambda\hat{y_i})  \right)
\end{align*}

A simpler approach for mean-only Gamma regression resorts to the deviance\footnote{Deviance is a measure of how well a proposed statistical model fits the data compared to a "perfect" model. This perfect model, known as the saturated model, has one parameter for every data point, meaning it fits the data perfectly but has no predictive power. Deviance is derived from the likelihood ratio test, since ratio of the likelihoods of the two models gives us an idea of how much worse our proposed model is compared to the perfect one: $\Lambda=\mathcal{L}_p/\mathcal{L}_s$. Deviance is formally defined as -2 times the log-likelihood ratio: $D=-2(\mathcal{L}_p -\mathcal{L}_s)$.
} of the Gamma distribution as loss function:
$$ \mathcal{L}_\Gamma = 2 \left( \frac{y-\hat{y}}{\hat{y}} - \log \frac{y}{\hat{y}} \right)$$
Unlike MSE and MAE, which penalizes absolute differences, Gamma deviance focuses on relative errors. It is invariant to scaling of the target variable. Its first term is the percent error relative to the prediction and the logarithmic term makes the loss function asymmetric (penalizes underpredictions much more heavily than overpredictions). As a result, the loss grows very quickly as your prediction approaches zero for a positive true value.

For the output layer, we can enforce a positive output using a positive activation function, such as softplus or exp ($e^z$), adding a small positive value $\varepsilon \approx 10^{-6}$ to avoid zero outputs. 





The Gamma GLM canonical link is the inverse link ($\eta = 1/\mu$). However, divisions by zero are problematic in computers, so a log link is more stable. In practice, you predict $\nu = \log \mu)$ and then output $\mu = e^\nu$.

\item
If your prediction target is a count (0, 1, 2...), i.e. the number of events, you can resort to a Poisson distribution:
$$ p(k) = \frac{\lambda^k e^{- \lambda}}{k!}$$
where the positive real number $\lambda \ge 0$ is equal to the expected value of the distribution (and also its variance).
$$ \log p(k) = k\log \lambda - \lambda - \log k!$$

In Poisson regression, the dependent variable $y$ is an observed count that follows the Poisson distribution whose rate $\lambda$ is determined as $\lambda = e^{w^\top x} = e^{\hat{y}}$ using a linear activation function.

Therefore, the log likelihood of our model is
$$ 
\log \mathcal{L} 
= \sum_{i=1}^n \left( y_i \hat{y_i} - e^{\hat{y_i}} - \log y_i! \right)
$$

An alternative loss function can be derived from the Poisson distribution calculating the deviance between the true count ($y$) and the predicted rate ($\hat{y})$:
$$ \mathcal{L}_{Poisson} = 2 \left( \hat{y} - y \log \hat{y} \right)$$

Since the Poisson loss function includes a $\log \hat{y}$ term, undefined for non-positive numbers, the model output must always be positive. An exponential is the canonical choice (the inverse link in the underlying GLM model). The softplus smooth approximation of the ReLU function might also be used.
It should also be noted that the target variable should not be scaled nor normalized, since it must represent raw counts.

%

\item
Tweedie distributions \cite{jorgensen1987} unify Gaussian ($p=0$), Poisson ($p=1$), Gamma ($p=2$), compound Poisson–Gamma, and inverse Gaussian ($p=3$) distributions. For positive continuous data with mass near zero, use $p \in (1,2]$, where $p=2$ corresponds to pure Gamma-like behavior. The loss per example is also derived from the distribution deviance:
$$ \mathcal{L}_{Tweedie} = \frac{2}{1-p} \left( y \hat{y}^{1-p} - \frac{y^{2-p}}{2-p} \right)$$
where the parameter $p$ can be tuned to match variance-mean relationship.  Gamma deviance aligns with variance scaling like $\sigma^2 \propto \mu^2$, log-MSE aligns with constant variance on the log scale, and Tweedie covers intermediate power laws ($\sigma^2 \propto \mu^p$).



As activation function for the output layer, you could use either softplus ($softplus(z)+\varepsilon$), exp ($e^z$), or even a square ($z^2 + \varepsilon$). Softplus is stable and smooth, exp strongly enforces positivity but can explode, and a square is less common because its gradients vanish near zero. However, a log link is usually recommended: just predict $\nu = \log \mu)$ using a linear output layer and then output $\mu = e^\nu$. 

\end{itemize}

MSE or MAE can be used in conjunction with ReLU or softplus activation functions when your data is roughly symmetric around the mean after a log transform. The logarithmic transformation handles skewed distributions (common in positive-only data) and can be used with log-normal-like heteroscedasticity, multiplicative noise, or many orders of magnitude in the output. The Gamma or Tweedie loss functions are statistically sound and suitable for skewed positive values. The Poisson loss can be used for predicting counts. The log link matches the multiplicative noise and skew of positive-value outcomes and should be the default choice for implementing Gamma/Poisson/Tweedie regression.

\subsection{Fat Tails}

The tails of the exponential family of probability distributions, which includes the Gaussian, Laplace, and Poisson distributions, are characterized by their tendency to decay exponentially, i.e. faster than polynomially, making them light-tailed distributions. Distributions that are not in the exponential family often exhibit heavy tails, where the decay is slower than exponential, such as power-law decay. These distributions, like the Student's t-distribution, Cauchy, and Pareto distributions, assign a much higher probability to extreme values.

The Gaussian assumption, which justifies the use of MSE, penalizes large errors more heavily because of the square error term and leads to models model that are especially sensitive to outliers or large deviations. The Laplace assumption lead to the use of MAE and is more robust to outliers, a better choice when you care about median-like behavior rather than mean-like behavior. The use of fat-tailed distributions can make models more robust to the presence of outliers.

The exponential family works well when data is well-behaved (light-tailed, symmetric), outliers are rare and not influential, and our main concern the average-case error in classification/regression. Fat-tailed distributions, with power-law instead of exponentially-decaying tails, are the natural choice when residuals/errors are heavy-tailed, extreme deviations matter and should not be washed out by excessive penalties, and robustness to outliers is sought without discarding tail information.

For example, let us assume a double symmetric Pareto distribution, with a sharp probability peak at $x=0$ and symmetric long probability tails derived from the Pareto distribution:
$$ p(x) = \frac{\alpha}{2} \frac{1}{(1+|x|)^{\alpha+1}}$$
where $\alpha$ is a shape parameter, known as the tail index, that controls the heaviness of the distribution tails. The double Pareto distribution 
exhibits power-law tails, a defining feature of scale-invariant properties (observed in many natural and economic phenomena). 

In a regression problem, the target variable $y$ is generated from our model prediction $\hat{y} = f(x)$ plus some fat-tailed (double Pareto) noise, $\epsilon$:
$$y = f(x) + \epsilon \quad \text{where} \quad \epsilon \sim \text{DoublePareto}(\alpha)$$
The noise $\epsilon$ is assumed to have a mean of 0 (the center of our symmetric double Pareto distribution).

This is equivalent to saying the likelihood of observing $y$ given $\hat{y}$ is:
$$p(y|\hat{y}) = \frac{\alpha}{2} \frac{1}{(1+|y-\hat{y}|)^{\alpha+1}}$$

Assuming a flat prior over $\hat{y}$, the Bayesian-optimal estimator is the maximum a posteriori (MAP) estimate:\footnote{MAP (maximum a posteriori) estimation and MLE (maximum likelihood estimation) are both statistical methods to find the most likely parameter values, but MAP includes prior knowledge, while MLE considers only the observed data. The MAP estimate is the mode of the posterior distribution. MAP finds the maximum of the posterior probability by combining the likelihood function with a prior distribution, while MLE finds the maximum of the likelihood function alone. MAP is a Bayesian approach, while MLE is a frequentist approach. MLE is a special case of MAP when the prior is uniform. }
$$ \hat{y}_{MAP} = \arg \max_{\hat{y}} p(y|\hat{y})$$
Taking logarithms:
$$ \log p(y|\hat{y}) = -(\alpha+1) \log (1+|y-\hat{y}|) + \log {(\alpha/2)} $$
Assuming that $\alpha$ can be considered constant, the MAP estimate minimizes the logarithmic loss
$$ \log \mathcal{L} = \log (1+|y-\hat{y}|)$$

The double Pareto loss behaves like MSE near zero, like MAE for moderate errors, and becomes even more robust in the tails due to its logarithmic growth, which resembles cross-entropy. Comparing its gradient with respect to the model output (against MSE and MAE) shows how double Pareto naturally down-weights outliers:

\begin{itemize}

\item
MSE: Linear gradient (large for big errors).
$$\frac{d}{d\hat{y}} \mathcal{L}_{MSE} = 2 (\hat{y} - y)$$  

\item
MAE: Constant gradient (more robust).
$$\frac{d}{d\hat{y}} \mathcal{L}_{MAE} = \text{sign}(\hat{y} -y)$$

\item
Log: Decaying gradient (robust and smooth).
$$ \frac{d}{d\hat{y}} \mathcal{L}_{log} = \frac{\text{sign}(\hat{y} - y)}{1 + |\hat{y} - y|}$$  

\end{itemize}

The double Pareto likelihood is symmetric and unimodal. Under a flat prior, the MAP estimate coincides with the mode of the likelihood.
This mirrors the Gaussian case, where the conditional mean is optimal under MSE, and the Laplace case, where the conditional median is optimal under MAE, but here, the conditional mode is optimal under double Pareto noise.\footnote{In a symmetric unimodal double Pareto distribution, the mode is the point of maximum density. The median, for a symmetric distribution centered at 0, is also 0. In asymmetric double Pareto distribution, with tails decaying at different rates, the mode, median, and mean are all distinct.
The mean exists only when the tail index $\alpha>1$ but is undefined when $\alpha \le 1$ because the integral diverges due to heavy tails. When extreme values dominate, the mean becomes unstable, and the MAP (mode) or median are preferred estimators.}

Let $Y$ be a random variable with conditional distribution given $X=x$. For any predictor $\hat{y}\in\mathbb{R}$, consider the conditional risk, i.e. the expected loss for a specific decision, given a particular observation:
$$R_{\log}(\hat{y}) = \mathbb{E}\!\left[\log (1+|Y-\hat{y}|)|X=x\right]$$
The Bayes estimator, or Bayes decision rule, minimizes the conditional risk for every possible observation:
$$
\frac{d}{d\hat{y}}R_{\log}(\hat{y}) = \frac{\text{sign}(\mathbb{E}[Y|x] - \hat{y})}{1+|\mathbb{E}[Y|x] - \hat{y}|}
$$
Setting the derivative to zero, we obtain that the conditional expectation the best possible prediction:
$$ 
\frac{\text{sign}(\mathbb{E}[Y|x] - \hat{y})}{1+|\mathbb{E}[Y|x] - \hat{y}|} = 0
$$
$$ 
\text{sign}(\mathbb{E}[Y|x] - \hat{y}) = 0
$$
$$
\frac{\mathbb{E}[Y|x] - \hat{y}}{|\mathbb{E}[Y|x] - \hat{y}|} = 0
$$
$$
\hat{y}^* = \mathbb{E}[Y|x].
$$
Logarithmic risk is minimized by the expectation and learning with the logarithmic loss function estimates \(\mathbb{E}[Y|X]\).

The Gaussian distribution underestimates tail risk and the mean is sensitive to outliers. The Laplace distribution captures moderate tails and the median is somewhat more robust to outliers. The double Pareto distribution explicitly models fat tails and is extremely robust to outliers, since the mode ignores tail mass.

Let us now model $\hat{y} = f(x)$ to find the optimal form of $f$. If we use a linear output layer $f(x)=w^\top x=\vec{w}\cdot\vec{x}$, is this form optimal under the double Pareto likelihood? 
The double Pareto loss is convex in $w^\top x$.
The true conditional expectation $\mathbb{E}[y|x]$ is linear in $x$ ($\mathbb{E}[y|x]=\hat{y}=w^\top x$) and, therefore, the Bayesian-optimal estimator under symmetric noise is also linear. A linear output layer is the optimal solution because the symmetry and convexity of the double Pareto likelihood preserve the optimality of linear predictors.

Since we are exploring loss–activation pairs, we can sketch how a double Pareto-inspired loss could also be paired with a non-saturating activation function, given that premature saturation would hide tail information. Potential candidates, for classification settings, would include:
\begin{itemize}
    \item 
    Softsign (similar to a sigmoid, but with polynomial tails consistent with Pareto-like decay):
    $$ f(z) = \frac{1}{1+|z|}$$

    \item
    Arctan (bounded, but with slower saturation than a sigmoid):
    $$ f(z) = \arctan(z)$$
\end{itemize}

Such pairs would be robust alternatives for heavy-tailed distributions: the loss functions discounts extreme deviations without ignoring them, whereas the activation allows gradients to flow even for large latent values (avoiding sigmoid vanishing gradient in the tails).

Fat tails beat exponential ones in situations with heavy-tailed leptokurtic distributions, which include financial and risk modeling, robust regression with outliers (e.g. spikes due to sensor faults), extreme event predictions (e.g. insurance claims, natural disasters, or network traffic spikes), and privacy-preserving or adversarial robust learning (i.e. when adversaries inject large perturbations).

\section{Conclusion}

Error (or loss) functions used for training deep learning models measure the discrepancy between the model's predictions and the true values. The choice of an appropriate error function is crucial for training a neural network effectively.

From a statistical point of view, the ``right'' loss function is simply the negative log-likelihood of the GLM corresponding to the assumed distribution of the response. That is the reason why MSE, MAE, BCE, and CCE are not just convenient, they are statistically principled. The same idea of negative log-likelihood minimization can be used to obtain a logarithmic loss function for regression under fat-tailed distributions, which should be the loss of choice for extreme value regression.

In classification problems, cross-entropy losses are the negative log-likelihoods of exponential family distributions (Bernoulli, multinomial) and are strictly proper scoring rules, uniquely minimized by the true probabilities. 

The activation function in the network output layer ensures that the network outputs lie in the correct domain (e.g. real values, probabilities, positive rates). The most suitable activation function can be justified as the inverse canonical link function of a GLM‑like likelihood model (for exponential distributions) or the Bayesian-optimal estimator under the assumed noise distribution (for fat-tailed distributions).

\appendix





\section{Activation functions}
\label{appendix-activation-functions}

\subsection{Logistic function}

The logistic or standard sigmoid function is defined as
$$\sigma(x) = \frac{1}{1 + e^{-x}}$$

The derivative of the logistic function has a remarkably simple and useful form:
$$\sigma'(x) = \frac{d}{dx}\sigma(x) = \sigma(x) \cdot (1 - \sigma(x))$$

This form is highly efficient for computation, especially in neural networks where the output $\sigma(x)$ is often already calculated during the forward pass. The derivative is a bell-shaped curve with a maximum value of $0.25$ at $x=0$, where $\sigma(0) = 0.5$ ($\sigma(x)(1 - \sigma(x))$ is maximized when $\sigma(x) = 1 - \sigma(x)$, or $\sigma(x) = 0.5$). As the input $|x|$ gets very large (positive or negative), $\sigma(x)$ approaches 1 or 0, making the derivative $\sigma(x)(1-\sigma(x))$ approach 0, which contributes to the vanishing gradient problem.

\noindent Step-by-step derivation:

\begin{itemize}
\item
Start with the logistic function:
$$\sigma(x) = \frac{1}{1 + e^{-x}} = (1 + e^{-x})^{-1}$$

\item 
Apply the chain rule: $u = 1 + e^{-x}$, so $\sigma(x) = u^{-1}$.
$$\frac{d}{dx}\sigma(x) = \frac{d}{du}(u^{-1}) \cdot \frac{d}{dx}(1 + e^{-x})$$

\item
Calculate the derivatives:
$$\frac{d}{du}(u^{-1}) = -1 \cdot u^{-2} = -u^{-2}$$   
$$\frac{d}{dx}(1 + e^{-x}) = 0 + \frac{d}{dx}(e^{-x}) = e^{-x} \cdot (-1) = -e^{-x}$$

\item
Substitute and simplify:
$$\frac{d}{dx}\sigma(x) = (-u^{-2}) \cdot (-e^{-x}) = \frac{e^{-x}}{u^2}$$
Substitute $u = 1 + e^{-x}$ back into the equation:
$$\sigma'(x) = \frac{e^{-x}}{(1 + e^{-x})^2}$$

\item
Express in terms of $\sigma(x)$:
$$\sigma'(x) = \frac{e^{-x}}{1 + e^{-x}} \cdot \frac{1}{1 + e^{-x}}$$
$$\frac{e^{-x}}{1 + e^{-x}} = \frac{1 + e^{-x} - 1}{1 + e^{-x}} = \frac{1 + e^{-x}}{1 + e^{-x}} - \frac{1}{1 + e^{-x}} = 1 - \sigma(x)$$
$$\sigma'(x) 
= (1 - \sigma(x)) \cdot \sigma(x)$$
\end{itemize}

\subsection{Hyperbolic tangent}

The hyperbolic tangent function is defined as
$$\tanh(x) = \frac{e^x- e^{-x}}{e^x + e^{-x}}$$

We can relate the hyperbolic tangent to the logistic function as follows:
\begin{align*}
\tanh(x)
& = \frac{e^x- e^{-x}}{e^x + e^{-x}} \\
& = \frac{e^x}{e^x} \frac{1- e^{-2x}}{1 + e^{-2x}} \\
& = \frac{2 - 1 - e^{-2x}}{1 + e^{-2x}} \\
& = \frac{2}{1 + e^{-2x}} - \frac{1 + e^{-2x}}{1 + e^{-2x}}\\
& = \frac{2}{1 + e^{-2x}} - 1\\
& = 2 \sigma(2x) - 1\\
\end{align*}

From the above relationship, we could also express the logistic function in terms of the hyperbolic tangent:
$$ \sigma(x) = \frac{1}{2} \left( \tanh \left( \frac{x}{2} \right) + 1 \right)$$

The derivative of the hyperbolic tangent function, $\tanh(x)$, is the hyperbolic secant squared function, $\text{sech}^2(x)$:
$$\frac{d}{dx} \tanh(x) = \text{sech}^2(x)$$

The derivative can be found by first expressing $\tanh(x)$ in terms of $\sinh(x)$ and $\cosh(x)$, and then applying the quotient rule and the fundamental hyperbolic identity:

\begin{itemize}

\item
Definition of the hyperbolic tangent and the derivatives of the hyperbolic sine and cosine functions:
$$\tanh(x) = \frac{\sinh(x)}{\cosh(x)}$$
$$\frac{d}{dx} \sinh(x) = \cosh(x)$$
$$\frac{d}{dx} \cosh(x) = \sinh(x)$$

\item
Applying the quotient rule, $\frac{d}{dx}\left(\frac{u}{v}\right) = \frac{u'v - uv'}{v^2}$, where $u = \sinh(x)$ and $v = \cosh(x)$:
$$\frac{d}{dx}\tanh(x) = \frac{d}{dx}\left(\frac{\sinh(x)}{\cosh(x)}\right)$$
$$\frac{d}{dx}\tanh(x) = \frac{(\cosh(x))(\cosh(x)) - (\sinh(x))(\sinh(x))}{(\cosh(x))^2}$$
$$\frac{d}{dx}\tanh(x) = \frac{\cosh^2(x) - \sinh^2(x)}{\cosh^2(x)}$$

\item
Given the fundamental hyperbolic identity, $\cosh^2(x) - \sinh^2(x) = 1$:
$$\frac{d}{dx}\tanh(x) = \frac{1}{\cosh^2(x)} = \text{sech}^2(x)$$
since the hyperbolic secant function is defined as $\text{sech}(x) = \frac{1}{\cosh(x)}$.

\end{itemize}

The derivative of $\tanh(x)$ can be also be expressed in terms of $\tanh(x)$ itself using the fundamental hyperbolic identity, the hyperbolic Pythagorean identity relating $\cosh(x)$ and $\sinh(x)$:
$$\cosh^2(x) - \sinh^2(x) = 1$$

Dividing every term in this identity by $\cosh^2(x)$, you get the desired relationship:
$$\frac{\cosh^2(x)}{\cosh^2(x)} - \frac{\sinh^2(x)}{\cosh^2(x)} = \frac{1}{\cosh^2(x)}$$
$$1 - \left(\frac{\sinh(x)}{\cosh(x)}\right)^2 = \left(\frac{1}{\cosh(x)}\right)^2$$
$$1 - \tanh^2(x) = \text{sech}^2(x)$$

Since $\frac{d}{dx}(\tanh(x)) = \text{sech}^2(x)$, we can substitute the expression above:

$$\frac{d}{dx} \tanh(x) = 1 - \tanh^2(x)$$

\subsection{Softmax function}

The softmax function, often used as an activation function in the output layer of a neural network for multi-class classification problems, is defined as:
$$y_i = \text{softmax}_i(\vec{x}) = \frac{e^{x_i}}{\sum_{k=1}^K e^{x_k}}$$
where $y_i$ is the $i$-th element of the output vector $\vec{y}$, $x_i$ is the $i$-th element of the weighted input vector $\vec{x}$, and the denominator is the sum over all $K$ classes.

The derivative of the softmax function is most easily expressed in terms of the Kronecker delta ($\delta_{ij}$), and it results in a matrix of values (a Jacobian matrix) because softmax is a vector-valued function. The derivative of the $i$-th component of the softmax output, $y_i$, with respect to the $j$-th component of the input, $x_j$, is
$$\frac{\partial y_i}{\partial x_j} = y_i (\delta_{ij} - y_j)$$
where $y_i$ and $y_j$ are the outputs of the softmax function, whereas 
$\delta_{ij}$ is the Kronecker delta ($1$ if $i = j$ and $0$ if $i \neq j$).

In the full derivation of the softmax derivative, we must consider two separate cases: when $i=j$ and when $i \neq j$.


\begin{itemize}

\item 
The derivative $\frac{\partial y_i}{\partial x_i}$with respect to the same index ($i = j$):
$$\frac{\partial}{\partial x_i}(e^{x_i}) = e^{x_i}$$
$$\frac{\partial}{\partial x_i}\left(\sum_{k=1}^N e^{x_k}\right) = e^{x_i}$$ 
(only the $k=i$ term in the sum depends on $x_i$)

$$\frac{\partial y_i}{\partial x_i} = \frac{(e^{x_i})(\sum_{k=1}^N e^{x_k}) - (e^{x_i})(e^{x_i})}{(\sum_{k=1}^N e^{x_k})^2}$$
$$\frac{\partial y_i}{\partial x_i} = \frac{e^{x_i}}{\sum_{k=1}^N e^{x_k}} \left( \frac{\sum_{k=1}^N e^{x_k} - e^{x_i}}{\sum_{k=1}^N e^{x_k}} \right)$$
$$\frac{\partial y_i}{\partial x_i} = y_i \left( 1 - \frac{e^{x_i}}{\sum_{k=1}^N e^{x_k}} \right) = y_i (1 - y_i)$$

\item
The derivative $\frac{\partial y_i}{\partial x_j}$ with respect to a different index ($i \neq j$):
$$\frac{\partial}{\partial x_j}(e^{x_i}) = 0$$ 
$$\frac{\partial}{\partial x_j}\left(\sum_{k=1}^N e^{x_k}\right) = e^{x_j}$$ 

$$\frac{\partial y_i}{\partial x_j} = \frac{(0)(\sum_{k=1}^N e^{x_k}) - (e^{x_i})(e^{x_j})}{(\sum_{k=1}^N e^{x_k})^2}$$
$$\frac{\partial y_i}{\partial x_j} = -\frac{e^{x_i} e^{x_j}}{(\sum_{k=1}^N e^{x_k})(\sum_{k=1}^N e^{x_k})}$$
$$\frac{\partial y_i}{\partial x_j} = -\left( \frac{e^{x_i}}{\sum_{k=1}^N e^{x_k}} \right) \left( \frac{e^{x_j}}{\sum_{k=1}^N e^{x_k}} \right)$$
$$\frac{\partial y_i}{\partial x_j} = -y_i y_j$$

\end{itemize}

Combining both cases:
$$\frac{\partial y_i}{\partial x_j} = \begin{cases} y_i(1 - y_i) & \text{if } i = j \\ -y_i y_j & \text{if } i \neq j \end{cases}$$
which can be written compactly using the Kronecker delta ($\delta_{ij}$) as
$$\frac{\partial y_i}{\partial x_j} = y_i (\delta_{ij} - y_j)$$

\subsection{Softplus function}

The softplus function \cite{dugas2000softplus}, or SmoothReLU function, is defined as 
$$ f(x) = \ln \left( 1 + e^x \right) $$

For large negative $x$, it is roughly $\ln 1$, just above $0$. For large positive $x$, it is roughly $\ln(e^x)$ ,just above $x$.

The derivative of the softplus function is the logistic function (the standard sigmoid function).

Given the softplus function, $f(x) = \ln(1 + e^x)$, the softplus derivative can be derived using the chain rule and simplified into the standard sigmoid function:

\begin{itemize}
\item
Apply the chain rule $\frac{d}{du}[\ln(u)] = \frac{1}{u} \cdot \frac{du}{dx}$, where $u = 1 + e^x$:
$$f'(x) = \frac{1}{1 + e^x} \cdot \frac{d}{dx}(1 + e^x)$$$$f'(x) = \frac{1}{1 + e^x} \cdot (0 + e^x)$$$$f'(x) = \frac{e^x}{1 + e^x}$$

\item
The form $\frac{e^x}{1 + e^x}$ is a common representation of the sigmoid function, $\sigma(x)$. To show the most common form of the sigmoid, $\frac{1}{1 + e^{-x}}$, just multiply both numerator and denominator by $e^{-x}$:
$$f'(x) = \frac{e^x}{1 + e^x} \cdot \frac{e^{-x}}{e^{-x}} = \frac{e^x \cdot e^{-x}}{e^{-x}(1 + e^x)} = \frac{1}{e^{-x} + 1}$$
\end{itemize}
Therefore, the derivative of softplus is:
$$\frac{d}{dx} (\text{softplus}(x)) = \sigma(x) = \frac{1}{1 + e^{-x}}$$


\subsection{Swish function}

The swish family function \cite{hendrycks2016gelu, elfwing2017silu, ramachandran2017swish} is defined as
$$\text{swish}_\beta (x) = x \sigma(\beta x) = \frac{x}{1 + e^{-\beta x}}$$
where $\beta$ can be constant or trained with the network parameters.

The swish family was designed to smoothly interpolate between a linear function and the ReLU function:

\begin{itemize}

\item
When $\beta=0$, the function is linear: 
$$\text{swish}_0 (x) = x/2$$

\item
When $\beta=1$, the function is the standard sigmoid or logistic: 
$$\text{swish}_1 (x) = \sigma(x)$$

\item
When $\beta \to \infty$, the function converges to the ReLU: 
$$\text{swish}_\infty (x) = \lim_{\beta \to \infty} \frac{x}{1 + e^{-\beta x}} = \text{ReLU(x)}$$
since for $x>0$, $e^{-\beta x} \to 0$ and, therefore, $\text{swish}_\infty (x) \to x$, whereas, for $x<0$, $e^{-\beta x} \to \infty$ and $\text{swish}_\infty (x) \to 0$.
\end{itemize}


The derivative of the swish function is given by
\begin{align*}
\text{swish}_\beta' (x) 
& = \sigma(\beta x) + \beta x \cdot \sigma'(\beta x) \\
& = \sigma(\beta x) + \beta x \cdot \sigma(\beta x) ( 1 - \sigma(\beta x)) \\
& = \beta x \cdot \sigma(\beta x) + \sigma(\beta x) ( 1 - \beta x \cdot \sigma(\beta x)) \\
& = \beta \cdot \text{swish}_\beta (x) + \sigma(\beta x) ( 1 - \beta \cdot  \text{swish}_\beta (x)) \\
\end{align*}

\subsection{ReLU function}

The rectifier or ReLU (rectified linear unit) activation function is defined as the non-negative part of its argument, i.e., the ramp function:
$$
    \text{ReLU}(x) = x^+ = \max \{ 0, x \} =  
        \begin{cases} 
            x & \text{if } x > 0 \\ 
            0 & \text{if } x \le 0
        \end{cases}
$$

The ReLU is analogous to half-wave rectification in electrical engineering and one of the most common activation functions for artificial neural networks.

Many variants of the ReLU function have been proposed in the literature:

\begin{itemize}

\item
Leaky ReLU \cite{maas2013leakyrelu} is a piecewise-linear variant that allows a small, positive gradient when the unit is inactive ($\alpha$ typically between 0.01 and 0.3):
$$
    \text{LReLU}(x) =
        \begin{cases} 
            x & \text{if } x > 0 \\ 
            \alpha x & \text{if } x \le 0
        \end{cases}
$$

\item
Parametric ReLU \cite{he2015prelu} makes the leaky ReLU $\alpha$ a learnable parameter. For $\alpha \le 1$:
$$ \text{PReLU}(x) = \max \{ x, \alpha x \}$$

\item
Concatenated ReLU \cite{shang2016crelu} preserves both positive and negative inputs by returning two values:
$$ \text{CReLU}(x) = [ \text{ReLU}(x), \text{ReLU}(-x)]$$

\end{itemize}

Apart from the softplus function, or SmoothReLU, additional smooth approximations to the rectifier function include:

\begin{itemize}

\item 
Exponential linear units, ELUs, which smoothly allow negative values \cite{clevert2016elu}:
$$
    \text{ELU}(x) =
        \begin{cases} 
            x & \text{if } x > 0 \\ 
            \alpha (e^x-1) & \text{if } x \le 0
        \end{cases}
$$
$$
    \text{ELU}'(x) =
        \begin{cases} 
            1 & \text{if } x > 0 \\ 
            \alpha e^x & \text{if } x \le 0
        \end{cases}
$$
ELU can be viewed as a smoothed version of a shifted ReLU (SReLU), which has the form $f(x)=\max \{ -\alpha, x\}$.

\item
Gaussian-error linear units, GELUs \cite{hendrycks2016gelu}:
$$GELU(x) = x \Phi (x)$$
$$GELU'(x) = x \Phi' (x) + \Phi (x)$$
where $\Phi (x)$ is the cumulative distribution function of the standard normal distribution $\mathcal{N}(0,1)$.
The swish function, or SiLU (sigmoid linear unit), is similar, also with a bump with negative derivative to the left of $x=0$, and computationally cheaper.

\item
The mish function \cite{misra2020mish} was obtained by experimenting with functions similar to swish  and exhibits a self-regularizing behavior attributed to a $\Delta(x)$ term in its first derivative:
$$\text{mish}(x) = x \tanh ( \text{softplus} (x))$$
$$\text{mish'}(x) = \Delta(x) \text{swish}_1 (x) + \frac{\text{mish}(x)}{x}$$
where $\Delta(x) = \text{sech}^2(\text{softplus}(x))$.

\item
Squareplus \cite{barron2021squareplus} is an algebraic softplus-like function
$$ \text{squareplus}(x) = \frac{x+\sqrt{x^2 + b}}{2}$$
$$ \text{squareplus}'(x) = \frac{1}{2} \left( \frac{x}{\sqrt{x^2 + b}} + 1 \right)$$
where the $b \ge 0$ hyperparameter determines the extent of the curved region near $x=0$. Similar to softplus, squareplus can be computed using only algebraic functions, making it suitable for computational efficiency and numerical stability.

\item
Extended exponential linear units, DELUs, are smoother within the neighborhood of zero and sharper for larger values \cite{ccatalbacs2023delu}:
$$
    \text{DELU}(x) =
        \begin{cases} 
            x & \text{if } x > x_c \\ 
            (e^{\alpha x}-1)/ \beta & \text{if } x \le x_c
        \end{cases}
$$
$$
    \text{DELU}'(x) =
        \begin{cases} 
            1 & \text{if } x > x_c \\ 
            (\alpha/\beta) e^{\alpha x} & \text{if } x \le x_c
        \end{cases}
$$
whose hyperparameters are typically set as $\alpha=1$, $\beta=2$, $x_c=1.25643$. The value of $x_c \ge 0$ results from imposing the continuity constraint for the chosen values of $\alpha$ and $\beta$:
$$ x_c = (e^{\alpha x_c}-1)/\beta$$

\end{itemize}

\section{Loss Functions}

Common loss functions are described in this Appendix, grouped by category. Well-known loss functions for classification problems (and matching probability distributions) include the following:

\begin{itemize}

\item
Cross-entropy loss (a.k.a. log loss): Standard for probabilistic classification, with special cases for binary (binary cross-entropy, BCE) and multi-class (categorical cross-entropy) classification.
$$\mathcal{L}_{CE} = H(y\|\hat{y}) = -\sum_k y_k \log \hat{y}_k$$  

\item
Focal loss \cite{lin2017, lin2020}: Down-weights easy examples, focuses on hard ones (common in imbalanced classification). Common for image segmentation.
$$\mathcal{L}_{FL} = - \sum_k y_k (1-\hat{y}_k)^\gamma \log \hat{y}_k$$
Focal loss adds a modulating factor $(1-\hat{y}_k)^\gamma$ to the cross entropy loss, with a tunable focusing parameter $\gamma \ge 0$.

\item
Hinge loss \cite{cortes1995}: Used in support vector machines (SVMs), with a geometrical justification. Margin-based, with a linear penalty when the margin is violated. For a binary classification problem using bipolar encoding for the target $y$ ($\{-1,+1\}$) and a classifier score $\hat{y}$:
$$\mathcal{L}_{HL} = \max \{ 0, 1 - y \hat{y} \}$$ 

The squared hinge loss is a variant of the hinge loss with squared penalty (quadratic penalty when the margin is violated):
$$\mathcal{L}_{SHL} = \left( \max \{ 0, 1 - y \hat{y} \} \right)^2$$  

For multi-class classification problems, the Crammer-Singer \cite{cramer2002} and Weston-Watkins \cite{weston1999} loss functions can be used:
$$\mathcal{L}_{CS} = \max \{0, 1 + \max_{k \ne y} \{\hat{y}_y - \hat{y}_k\}\}$$ 
$$\mathcal{L}_{WW} = \sum_{k \ne y} \max \{ 0, 1 + \hat{y}_y \hat{y}_k \}$$

\item
Kullback–Leibler divergence, or KL loss \cite{kl1951}, also known as the relative entropy of $y$ with respect to $\hat{y}$: 
$$\mathcal{L}_{KL} = D_{KL}(y\|\hat{y}) = \sum_k y_k \log \frac{y_k}{\hat{y}_k}$$

The KL divergence is an f-divergence measure between true and predicted distributions.  

Since $y$ is usually represented using one-hot encoding, it simplifies to the cross-entropy loss, given that $D_{KL}(y\|\hat{y}) = H(y\|\hat{y}) - H(y)$.

\item
Hellinger divergence, a.k.a. Hellinger distance \cite{hellinger1909}: Another f-divergence that quantifies the similarity between two probability distributions:
$$\mathcal{L}_{HD} = D_{H}(y\|\hat{y}) = \frac{1}{\sqrt{2}} \sqrt{\sum_k \left( \sqrt{y_k} - \sqrt{\hat{y_k}} \right)^2 }$$
which is directly related to the Euclidean norm of the difference of the square root vectors. Sometimes the factor $1/\sqrt{2}$ is omitted, in which case the Hellinger distance ranges from zero to the square root of two.

\item
Total variation distance, a.k.a. statistical distance, statistical difference,or variational distance: Another f-divergence that provides a statistical distance between probability distributions.
$$\mathcal{L}_{TVD} = D_{TVD}(y\|\hat{y}) = \sum_k | y_k - \hat{y_k} |$$
i.e. half of the L1 distance between the probability functions on discrete domains.

The three f-divergences \cite{renyi1961} above are related:
$$ D_{TVD}(y\|\hat{y}) \le \sqrt{ \frac{1}{2} D_{KL}(y\|\hat{y})} $$
$$ D_{H}^2(y\|\hat{y}) \le D_{TVD}(y\|\hat{y}) \le \sqrt{2} D_{H}(y\|\hat{y})$$
i.e. inequalities that follow immediately from the relationship between the 1-norm and the 2-norm.

\item
Jensen–Shannon divergence, also known as information radius (IRad) or total divergence to the average: Yet another f-divergence, a symmetrized and smoothed version Kullback–Leibler divergence that always has a finite value.
$$\mathcal{L}_{JS} = D_{JS}(y\|\hat{y}) = \frac{D_{KL}(y\|m)+D_{KL}(\hat{y}\|m)}{2}$$
where $m=(y+\hat{y})/2$ is a mixture of $y$ and $\hat{y}$ (i.e. the pointwise mean of $y$ and $\hat{y}$ probabilities). The square root of the Jensen–Shannon divergence is a metric often referred to as Jensen–Shannon distance, $\sqrt{D_{JS}(y\|\hat{y})}$.

\item
Wasserstein Distance \cite{kantorovich1942, vaserstein1969}, a.k.a. the Earth mover’s distance or the optimal transport distance: Another similarity metric between two probability distributions. It can be interpreted as the minimum energy cost of moving and transforming a pile of dirt in the shape of one probability distribution to the shape of the other distribution. The cost is calculated as the product of the amount of probability mass being moved and the distance it is being moved. 
$$ \delta_{0} = 0 $$
$$ \delta_{k+1} = \delta_k + y_k - \hat{y}_k $$
$$\mathcal{L}_{W_1} = W_1(y,\hat{y}) = \sum_k  | \delta_k |$$
In general, for any Minkowski distance:
$$\mathcal{L}_{W_p} = W_p(y,\hat{y}) = \left( \sum_k  || \delta_k ||^p \right) ^{1/p}$$

\item
Bhattacharyya distance \cite{bhattacharyya1946}: Not a metric, actually, since it does not satisfy the triange inequality.
$$\mathcal{L}_{B} = D_B(y,\hat{y}) = -\log BC(y,\hat{y})$$
where
$$ BC(y,\hat{y}) = \sum_k \sqrt{ y_k \hat{y_k} }$$
is the Bhattacharyya coefficient for discrete probability distributions.

\item
Renyi divergence, or $\alpha$-divergence \cite{renyi1961}: A spectrum of divergence measures generalize the KL divergence. Its order $\alpha$ controls how sensitive you want to be to rare differences.
$$\mathcal{L}_{\alpha} = D_{\alpha}(y\|\hat{y}) = \frac{1}{\alpha-1} \log \sum_k \frac{y_k^{\alpha}}{\hat{y}_k^{\alpha-1}}$$

For special values, we can define the Rényi divergence by taking a limit:
\begin{itemize}
\item 
As $\alpha \to 0$, $D_0$ measures how much $\hat{y}$ supports $y$, checking whether $\hat{y}$ assigns probability to all points where $y$ is nonzero.
$$ D_{0}(y\|\hat{y}) = - \log \sum_k \hat{y} \cdot 1_{y_k>0}$$
where $1_{y_k>0}$ is an indicator function that is 1 if $y_k>0$, and 0 otherwise.

\item 
As $\alpha \to 1$, the Rényi divergence converges to the Kullback–Leibler (KL) divergence:
$$D_{1}(y\|\hat{y}) = D_{KL}(y\|\hat{y}) = \sum_k y_k \log \frac{y_k}{\hat{y}_k}$$

\item 
As $\alpha \to \infty$, $D_{\infty}$ becomes the max-divergence, focusing only on the largest discrepancy between the two distributions:
$$ D_{\infty}(y\|\hat{y}) = \log \max_k \frac{y_k}{\hat{y}_k}$$

\item 
When $\alpha = 1/2$, $D_{1/2}$ is twice the Bhattacharyya divergence:
$$D_{2}(y\|\hat{y}) = 2 D_{B}(y\|\hat{y}) = -2 \log \sqrt{ y_k \hat{y_k} }$$

\item 
When $\alpha = 2$, $D_{2}$ is the logarithm of the expected ratio of probabilities:
$$D_{2}(y\|\hat{y}) =  \log \left( y_k \frac{y_k}{\hat{y}_k} \right) = \log \left\langle \frac{y_k}{\hat{y}_k} \right\rangle$$

\end{itemize}

The Rényi divergence is indeed a divergence, i.e. it is greater than or equal to zero, and zero only when both distributions are the same. For any given pair of distributions, the Rényi divergence is nondecreasing as a function of its order $\alpha$. 

\end{itemize}

For regression problems, a wide variety of loss functions have been proposed:

\begin{table*}[t!]
\begin{tabular}{SlScScSc} 
\toprule[1.5pt]
Loss function & Small errors& Large errors & Derivative $\partial \mathcal{L} / {\partial \hat{y}_i} $ \\ 
\midrule[1pt]
MSE
  & $\epsilon^2$
  & $\epsilon^2$
  & $\epsilon$ 
  \\
MAE
  & $\epsilon$
  & $\epsilon$
  & $1$
  \\
Huber
  & $\epsilon^2$
  & $\epsilon$
  & $1$
  \\
Log-cosh
  & $\epsilon^2$
  & $\epsilon$
  & $1$
  \\
Hinge
  & $0$
  & $\epsilon$
  & $1$
  \\
Logarithmic
  & $\log \epsilon$
  & $\log \epsilon$
  & $1/\epsilon$
  \\
Cauchy
  & $\log \epsilon^2$
  & $\log \epsilon^2$
  & $1/\epsilon$
  \\
Student
  & $\log \epsilon^2$
  & $\log \epsilon^2$
  & $1/\epsilon$
  \\
Fair
  & $\epsilon$
  & $\epsilon$
  & $1$
  \\
Tukey
  & $\epsilon^2$
  & $1$
  & $0$
  \\  
Pinball/quantile
  & $\epsilon$
  & $\epsilon$
  & $1$
  \\  
\bottomrule[1.5pt]
\end{tabular}
\caption{Regression loss functions and their asymptotic behavior with small and large errors.}
\label{regression}
\end{table*}

\begin{itemize}

\item
Mean Squared Error, MSE \cite{gauss1809}: Highly sensitive to outliers, but smooth and differentiable. Equivalent to MLE under Gaussian noise with constant variance.
$$\mathcal{L}_{MSE} = \frac{1}{n}\sum_i (y_i - \hat{y}_i)^2$$  
$$\frac{\partial \mathcal{L}_{MSE}}{\partial \hat{y}_i} = -2(y_i - \hat{y}_i)$$

\item
Mean Absolute Error, MAE \cite{laplace1774}, a.k.a. L1 Loss: More robust to outliers, but non-differentiable at 0. MLE under Laplace (double exponential) noise, leads to median regression.
$$\mathcal{L}_{MAE} = \frac{1}{n}\sum_i |y_i - \hat{y}_i|$$  
$$\frac{\partial \mathcal{L}_{MAE}}{\partial \hat{y}_i} = -\text{sign}(y_i - \hat{y}_i)$$   

\item
Huber loss \cite{huber1964}: Quadratic near 0, linear for large residuals. Balances MSE for small errors and MAE for large errors. Robust, differentiable, widely used in robust regression.
  $$
  \mathcal{L}_{Huber} = 
  \begin{cases}
    \tfrac{1}{2}(y - \hat{y})^2 & |y - \hat{y}| \leq \delta \\
    \delta(|y - \hat{y}| - \tfrac{1}{2}\delta) & |y - \hat{y}| > \delta
  \end{cases}
  $$
  $$
  \frac{\partial \mathcal{L}_{Huber}}{\partial \hat{y}_i} = 
  \begin{cases}
    -(y_i - \hat{y}_i) & |y_i - \hat{y}_i| \leq \delta \\
    -\delta \,\text{sign}(y_i - \hat{y}_i) & |y_i - \hat{y}_i| > \delta
  \end{cases}
  $$

\item
Log-cosh loss: Smooth approximation to MAE, less harsh than MSE, with bounded gradient, shares similarities with Huber loss but is infinitely differentiable.
$$\mathcal{L}_{LCL} = \sum_i \log\cosh(y_i - \hat{y}_i)$$  
$$\frac{\partial \mathcal{L}_{LCL}}{\partial \hat{y}_i} = -\tanh(y_i - \hat{y}_i)$$

\item
Hinge loss for regression, a.k.a. $\epsilon$-insensitive loss \cite{vapnik1995}: Robust to small noise, focuses on large deviations.
Support Vector Regression (SVR) generalizes hinge loss to continuous targets.   Unlike MSE and MAE, SVR ignores small noise: errors smaller than the $\epsilon$ margin of tolerance are ignored (SVR has a dead zone of zero penalty, which makes it less sensitive to tiny fluctuations). Errors larger than  $\epsilon$ are penalized linearly, as in MAE. 
$$ \mathcal{L}_{SVR} = \mathcal{L}_{\epsilon}= \max(0, |y - \hat{y}| - \epsilon) $$
$$ 
\frac{\partial \mathcal{L}_{SVR}}{\partial \hat{y}} 
= \begin{cases} 0 & |y - \hat{y}| \leq \epsilon \\ -\text{sign}(y - \hat{y}) & |y - \hat{y}| > \epsilon \end{cases} 
$$
The hinge loss can be justified geometrically, connecting regression to margin-based learning: SVR finds a function such that most data points lie within an $\epsilon$-tube, while keeping the function “flat”. However, the hinge loss is not a proper scoring rule (the forecasted distribution does not match the distribution of the observation, i.e. the true distribution). 
Variations of the hinge loss include:

- $\epsilon$-insensitive squared loss (MSE-like, with quadratic penalty outside the tube):
$$ 
\mathcal{L}_\epsilon^{(2)} = \big(\max\{0, |y - \hat{y}| - \epsilon\}\big)^2 $$

- Huberized $\epsilon$-insensitive loss (Huber-like, more numerically stable than sharp $\epsilon$-insensitive): Flat inside the tube, quadratic near the boundary, linear far out.
$$ \mathcal{L}_{\epsilon,\delta} = 
\begin{cases} 
0 & |y - \hat{y}| \leq \epsilon \\ 
\tfrac{1}{2\delta}(|y - \hat{y}| - \epsilon)^2 & \epsilon < |y - \hat{y}| \leq \epsilon + \delta \\ 
|y - \hat{y}| - \epsilon - \tfrac{\delta}{2} & |y - \hat{y}| > \epsilon + \delta 
\end{cases} 
$$

- Pinball (quantile) loss in SVR: Useful for predicting intervals. If you set $\epsilon=0$ and choose asymmetric penalties, SVR becomes quantile regression.
$$ \mathcal{L}_\tau = \begin{cases} \tau (y - \hat{y}) & y \geq \hat{y} \\ (1-\tau)(\hat{y} - y) & y < \hat{y} \end{cases} $$



\item
Logarithmic loss for regression (i.e. Pareto loss): Heavy-tailed, robust to outliers, decaying gradient. MLE under double Pareto noise. In probabilistic regression, the log score is a proper scoring rule, ensuring calibrated predictive distributions.
$$
\mathcal{L}_{log} 
= \sum_i \log\left(1 + |y_i - \hat{y}_i|\right)
$$  
$$
\frac{\partial \mathcal{L}_{log}}{\partial \hat{y}_i} 
= \frac{\text{sign}(\hat{y}_i - y_i)}{1 + |\hat{y}_i - y_i|}
$$

\item
Cauchy loss: Heavy-tailed, robust to outliers. MLE under Cauchy noise.
$$
\mathcal{L}_{Cauchy} 
= \sum_i \log\left(1 + \frac{(y_i - \hat{y}_i)^2}{c^2}\right)
$$  
$$
\frac{\partial \mathcal{L}_{Cauchy}}{\partial \hat{y}_i} 
= -\frac{2(y_i - \hat{y}_i)}{c^2 + (y_i - \hat{y}_i)^2}
$$

\item
Student-t loss: Heavy-tailed, robust to outliers. MLE under Student-t noise.
$$
\mathcal{L}_{Student} 
= \sum_i \frac{\nu+1}{2} \log\left(1 + \frac{(y_i - \hat{y}_i)^2}{\nu \sigma^2}\right)
$$  
$$
\frac{\partial \mathcal{L}_{Student}}{\partial \hat{y}_i} 
= - \frac{(\nu+1)(y_i - \hat{y}_i)}{(y_i - \hat{y}_i)^2 +\sigma^2\nu}
$$

\item
Fair loss \cite{fair1974}: Smooth, large residuals with bounded influence, less aggressive than Cauchy.
$$
\mathcal{L}_{Fair} 
= c^2\left(\frac{|y - \hat{y}|}{c} - \log\left(1 + \frac{|y - \hat{y}|}{c}\right)\right)
$$
$$
\frac{\partial \mathcal{L}_{fair} }{\partial \hat{y}} 
= -\frac{y - \hat{y}}{1 + |y - \hat{y}|/c}
$$

\item 
Tukey's biweight loss \cite{tukey1977}: Bounded influence, completely ignores very large residuals beyond theshold $c$. Extremely robust, but non-convex. Used for outlier rejection.
  $$
  \mathcal{L}_{Tukey} = 
  \begin{cases}
    \frac{c^2}{6}\left[1 - \left(1 - \left(\frac{y - \hat{y}}{c}\right)^2\right)^3\right] & |y - \hat{y}| \leq c \\
    \frac{c^2}{6} & |y - \hat{y}| > c
  \end{cases}
  $$
For small error values, Tukey's biweight loss function can be closely approximated by a simple quadratic polynomial, behaving much like the standard squared error loss: $\mathcal{L}_{Tukey} \approx {\epsilon^2}/{2} - {\epsilon^4}/{2c^2}$ from the Taylor series expansion of the loss function around $\epsilon = 0$.

\item
Pinball loss, a.k.a. quantile loss \cite{koenker1978}:  For quantile regression, asymmetric penalty depending on quantile $\tau$. MLE under asymmetric Laplace distribution  computes conditional quantiles instead of mean. Useful for heteroskedastic or skewed data.
$$\mathcal{L}_{pinball} = \sum_i \max\{\tau(y_i - \hat{y}_i), (1-\tau)(y_i - \hat{y}_i)\}$$
The function gets its name from its characteristic V shape, which is asymmetrical when $\tau$ is not 0.5: for underprediction ($y_i \ge \hat{y}_i$), the loss is weighted by $\tau$; for overprediction ($y_i < \hat{y}_i$), the loss is weighted by $1-\tau$. This weighting encourages the model to make predictions that are more likely to be above ($\tau>0.5$) or below ($\tau<0.5$) the actual value, depending on the chosen quantile.


\end{itemize}




\nocite{*}
\bibliographystyle{acm}
\bibliography{references}

\end{document}